\documentclass{article}

\usepackage[dvipsnames]{xcolor}

\def\eg{\emph{e.g.}}
\def\ie{\emph{i.e.}}

\definecolor{turquoise}{cmyk}{0.65,0,0.1,0.3}
\definecolor{purple}{rgb}{0.65,0,0.65}
\definecolor{dark_green}{rgb}{0, 0.5, 0}
\definecolor{orange}{rgb}{0.8, 0.6, 0.2}
\definecolor{red}{rgb}{0.8, 0.2, 0.2}
\definecolor{darkred}{rgb}{0.6, 0.1, 0.05}
\definecolor{blueish}{rgb}{0.0, 0.3, .6}
\definecolor{light_gray}{rgb}{0.7, 0.7, .7}
\definecolor{pink}{rgb}{1, 0, 1}
\definecolor{greyblue}{rgb}{0.25, 0.25, 1}
\definecolor{pastelgreen}{rgb}{0.47, 0.87, 0.47}
\definecolor{teagreen}{rgb}{0.82, 0.94, 0.75}
\definecolor{cobalt}{rgb}{0.0, 0.28, 0.67}
\definecolor{lightskyblue}{rgb}{0.53, 0.81, 0.98}
\definecolor{paleblue}{rgb}{0.69, 0.93, 0.93}
\definecolor{palecornflowerblue}{rgb}{0.67, 0.8, 0.94}

\definecolor{airforceblue}{rgb}{0.36, 0.54, 0.66}
\definecolor{ao(english)}{rgb}{0.0, 0.5, 0.0}
\definecolor{azure(colorwheel)}{rgb}{0.0, 0.5, 1.0}
\definecolor{crimson}{rgb}{0.86, 0.08, 0.24}
\definecolor{darkcerulean}{rgb}{0.03, 0.27, 0.49}
\definecolor{cobalt}{rgb}{0.0, 0.28, 0.67}
\definecolor{rosegold}{rgb}{0.72, 0.43, 0.47}
\definecolor{orange-red}{rgb}{1.0, 0.27, 0.0}
\definecolor{mountainmeadow}{rgb}{0.19, 0.73, 0.56}
\definecolor{malachite}{rgb}{0.04, 0.85, 0.32}
\definecolor{darkblue}{rgb}{0.0, 0.0, 0.55}
\definecolor{customblue}{rgb}{0.2, 0.35, 0.8}

\definecolor{graycell}{gray}{0.85} % Define the gray color used for the cells

\definecolor{Red}{rgb}{0.6,0,0}
\definecolor{Blue}{rgb}{0,0,1}
\definecolor{Green}{rgb}{0,0.4,0.7}
\definecolor{airforceblue}{rgb}{0.36, 0.54, 0.66}
\definecolor{ao(english)}{rgb}{0.0, 0.5, 0.0}
\definecolor{azure(colorwheel)}{rgb}{0.0, 0.5, 1.0}
\definecolor{darkcerulean}{rgb}{0.03, 0.27, 0.49}
\definecolor{cobalt}{rgb}{0.0, 0.28, 0.67}
\definecolor{rosegold}{rgb}{0.72, 0.43, 0.47}
\definecolor{orange-red}{rgb}{1.0, 0.27, 0.0}
\definecolor{mountainmeadow}{rgb}{0.19, 0.73, 0.56}
\definecolor{malachite}{rgb}{0.04, 0.85, 0.32}
\definecolor{darkblue}{rgb}{0.0, 0.0, 0.55}

\definecolor{customcolor}{gray}{0.}
\definecolor{gg}{gray}{0.9}
\definecolor{tg}{gray}{0.6}
\usepackage{amsmath,amsfonts,bm}

\newcommand{\loss}[1]{\mathcal{L}_\text{#1}}
\newcommand{\expect}{\mathbb{E}}

\def\eqref#1{equation~\ref{#1}}
\def\1{\bm{1}}

\DeclareMathAlphabet{\mathsfit}{\encodingdefault}{\sfdefault}{m}{sl}
\SetMathAlphabet{\mathsfit}{bold}{\encodingdefault}{\sfdefault}{bx}{n}

\definecolor{mygreen}{rgb}{0.678, 1, 0.184}
\definecolor{myblue}{rgb}{0.678, 0.847, 1}

\PassOptionsToPackage{square}{natbib} % ywlee
    \usepackage[final]{template/neurips2024/neurips_2024}
    \setcitestyle{numbers} % ywlee

\usepackage{microtype}
\usepackage[pdftex]{graphicx}
\usepackage{wrapfig}
\usepackage{tabularx}
\usepackage{booktabs} % for professional tables
\usepackage{float}

\usepackage{adjustbox}
\usepackage{multirow}
\usepackage{subcaption}
\usepackage{pifont}
\usepackage{array}

\usepackage[pagebackref,breaklinks,colorlinks]{hyperref}
\usepackage[capitalize,noabbrev,nameinlink]{cleveref}
\usepackage{xcolor}
\usepackage{colortbl}
\usepackage{enumitem}

\usepackage{hyperref}
\hypersetup{colorlinks=true}
\hypersetup{linktoc=all}
\hypersetup{citecolor=teal}
\hypersetup{linkcolor=crimson}
\hypersetup{urlcolor=darkblue}
\usepackage[all]{hypcap}

\crefname{figure}{Fig.}{Figs.}
\Crefname{figure}{Fig.}{Figs.}
\crefname{table}{Tab.}{Tabs.}
\Crefname{table}{Tab.}{Tabs.}
\crefname{section}{Sec.}{Secs.}
\Crefname{section}{Sec.}{Secs.}
\crefname{appendix}{App.}{Apps.}
\Crefname{appendix}{App.}{Apps.}

\begin{document}

% \title{\includegraphics[width=0.08\linewidth]{figures/koala.pdf}KOALA: Self-Attention Matters in Knowledge Distillation of Latent Diffusion Models for Memory-Efficient and Fast Image Synthesis}
% \title{KOALA: Self-Attention Matters in Knowledge Distillation of Latent Diffusion Models for Memory-Efficient and Fast Image Synthesis}

% \title{KOALA: An empirical study on memory-Efficient and Fast diffusion based Image synthesis}
\title{
KOALA: Empirical Lessons Toward Memory-Efficient and Fast Diffusion Models for Text-to-Image Synthesis
}

\newcommand*{\affaddr}[1]{#1} % No op here. Customize it for different styles.
\newcommand*{\affmark}[1][*]{\textsuperscript{#1}}
\newcommand*{\email}[1]{\texttt{#1}}

\author{%
Youngwan Lee\affmark[1,2]~~~Kwanyong Park\affmark[1]~~~Yoorhim Cho\affmark[3]~~~Yong-Ju Lee\affmark[1]~~~Sung Ju Hwang\affmark[2,4]\\ 
\vspace{-0.1in}
\\
\affaddr{\affmark[1]Electronics and Telecommunications Research Institute~(ETRI), South Korea}\\
\affaddr{\affmark[2]Korea Advanced Institute of Science and Technology~(KAIST), South Korea}\\
\affaddr{\affmark[3]Sungkyunkwan University, South Korea}\\
\affaddr{\affmark[4]DeepAuto.ai, South Korea}\\
\footnotesize{project page: \url{https://youngwanlee.github.io/KOALA/}}
}

\maketitle

\begin{abstract}
% Motviation: 
% 1) Recent T2I mdoel's higher inference cost 
% 2) challenges in reproduction due to the restricted data
% Our goal:
% 1) reducing inference cost by proposing efficient T2I with KD
% 2) exploring how far we push the bar

As text-to-image (T2I) synthesis models increase in size, they demand higher inference costs due to the need for more expensive GPUs with larger memory, which makes it challenging to reproduce these models in addition to the restricted access to training datasets. Our study aims to reduce these inference costs and explores how far the generative capabilities of T2I models can be extended using only publicly available datasets and open-source models. To this end, by using the de facto standard text-to-image model, Stable Diffusion XL~(SDXL), we present three key practices in building an efficient T2I model: (1) \textbf{Knowledge distillation}: we explore how to effectively distill the generation capability of SDXL into an efficient U-Net and find that self-attention is the most crucial part. (2) \textbf{Data}: despite fewer samples, high-resolution images with rich captions are more crucial than a larger number of low-resolution images with short captions. (3) \textbf{Teacher}: Step-distilled Teacher allows T2I models to reduce the noising steps. Based on these findings, we build two types of efficient text-to-image models, called KOALA-Turbo~\&-Lightning, with two compact U-Nets~(1B \& 700M), reducing the model size up to 54\% and 69\% of the SDXL U-Net. In particular, the KOALA-Lightning-700M is $4\times$ faster than SDXL while still maintaining satisfactory generation quality. 
Moreover, unlike SDXL, our KOALA models can generate 1024px high-resolution images on consumer-grade GPUs with 8GB of VRAMs~(3060Ti). We believe that our KOALA models will have a significant practical impact, serving as cost-effective alternatives to SDXL for academic researchers and general users in resource-constrained environments.

\end{abstract}

\section{Introduction}
\label{sec:intro}

% Background: SDXL의 등장 (SDM건너뛰고 이제 바로 SDXL부터)
% Background: 
% 1) SDXL's higher inference cost <- de facto로 강조해서 SDXL의 경량화 시도 필요성 제시
% 2) challenges in reproduction due to the restricted data
Since Stable Diffusion XL~\cite{podell2023sdxl} (SDXL) has become the de facto standard model for text-to-image~(T2I) synthesis due to its ability to generate high-resolution images and its open-source nature, many models for specific downstream tasks~\cite{chen2024textdiffuser,shi2023mvdream,blattmann2023svd,zhang2023i2vgen,yang2024mastering} now leverage SDXL as their backbone.
However, the model's massive computational demands and large size necessitate expensive hardware, thus incurring significant costs in terms of training and inference.
Moreover, recent T2I works~\cite{chen2023pixartalpha, chen2024pixartsigma, sauer2023sdxl-turbo, dalle-3} do not release their training datasets, making it challenging for the open-source community to reproduce their performance due to internal or commercial restrictions.

% Motivation (problems to solve, limitations of related works)
To alleviate the computation burden, previous works have resorted to quantization~\cite{shang2023q-diff}, hardware-aware optimization~\cite{chen2023speed}, denoising step reduction~\cite{salimans2022progressive,meng2023distillation, sauer2023sdxl-turbo,lin2024lightning}, and architectural model optimization~\cite{kim2023bksdm}.
In particular, the denoising step reduction~(\ie, step-distillation) and architectural model compression methods adopt the knowledge distillation~(KD) scheme~\cite{hinton2015outkd,heo2019featkd2} by allowing the model to mimic the output of the SDMs as a teacher model.
% The step-distillation methods~\cite{salimans2022progressive,meng2023distillation,li2023snapfusion} allow the denoised latent of the diffusion model in the early denoising steps to mimic the output in the later denoising steps of the teacher model.
% As an orthogonal work for the architectural model compression, BK-SDM~\cite{kim2023bksdm} exploits KD when compressing the most heavy-weight part, U-Net, in SDM-v1.4~\cite{sdm-v1.4}. 
For the architectural model compression, BK-SDM~\cite{kim2023bksdm} exploits KD when compressing the most heavy-weight part, U-Net, in SDM-v1.4~\cite{sdm-v1.4}. 
BK-SDM builds a compressed U-Net by simply removing some blocks, which allows the compressed U-Net to mimic the last features at each stage and the predicted noise from the teacher model during the pre-training phase.
However, the compression method proposed by BK-SDM achieves a limited compression rate~(33\% in~\cref{tab:arch}) when applied to the larger SDXL than SDM-v1.4, and the strategy for feature distillation in U-Net has \textit{not yet been fully explored}.

\begin{figure*}[t!]
\centering
    \includegraphics[width=0.97\textwidth]{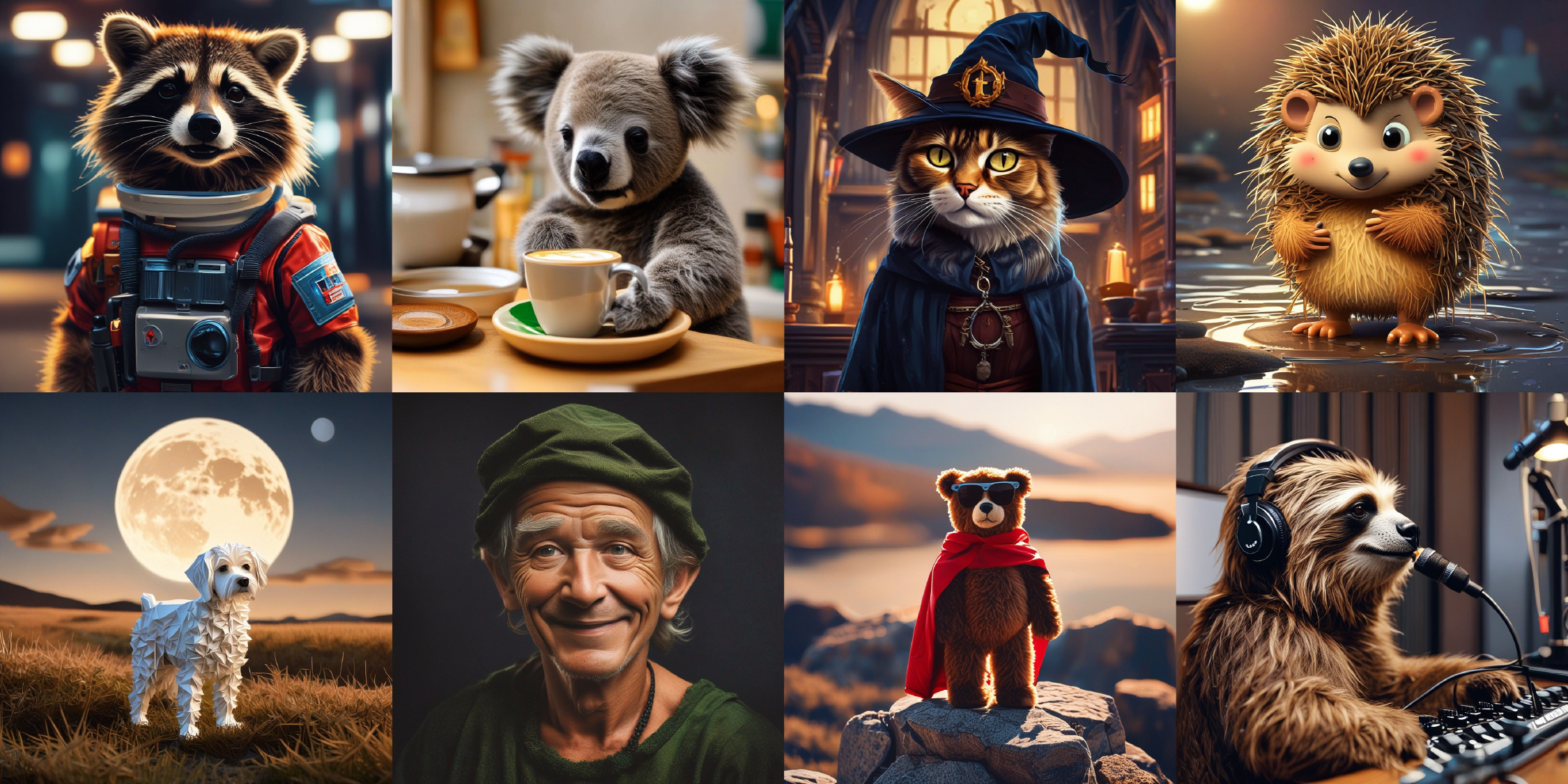}
\vspace{-0.1cm}
% \caption{\textbf{Qualiatative comparison with SDM-v2.0 and SDXL-Base-1.0.} 
\caption{\textbf{Samples by KOALA-Lightning-700M} with $\mathbf{1024^2}$ resolution and 10 denoising steps, generated in 0.65 seconds on NVIDIA 4090 GPU. The prompts and more qualitative comparisons are illustrated in~\cref{sec:app_samples}.
} \label{fig:teaser}
\vspace{-0.4cm}
\end{figure*}

% \begin{figure*}[ht]
% \centering
% \includegraphics[width=\textwidth]{figures/fig_teaser.pdf}
% % \vspace{-0.3cm}
% \caption{\textbf{Generated samples by our KOALA-700M trained by the proposed knowledge-distillation approach with SDXL~\cite{podell2023sdxl}}. With the following settings: FP-16 precision, $1024\times1024$ resolution, and 25 denoising steps with Euler discrete scheduler~\cite{karras2022elucidating} same as the huggingface's SDXL-Base-1.0 model~\cite{sdxl_hf}, the inference time is 1.4 seconds on an NVIDIA 4090~ (24GB) GPU, which is over $2\times$ faster than SDXL-Base-1.0~(3.3s) while reducing the U-Net model size by 69\%.
% }
% \label{fig:teaser}
% % \vspace{-0.3cm}
% \end{figure*}

% \begin{multicols}{2}
% % your text
% \begin{figure*}[htb!]
% \centering
%     \includegraphics[width=\textwidth]{figures/teaser.pdf}
%     \caption{Your Caption}
% \end{figure*}
% \label{fig:teaser}
% % more text
% \end{multicols}

% Our method
% In this work, our goal is to build an efficient T2I model by exploring how far we can push the performance using only open-source data and models alone.
In this work, our goal is to build an efficient T2I model based on \textbf{k}n\textbf{o}wledge distill\textbf{a}tion in the \textbf{la}tent diffusion model~(KOALA) by exploring how far we can push the performance using only open-source data and models alone.
% To this end, we explore three ways; 1) we design two efficient U-Nets, \textbf{KOALA-1B} and \textbf{KOALA-700M}, using not only block removal but also \textit{layer-wise trimming} to reduce the model size of the SDXL U-Net by up to 54\% and 69\%~(vs. BK's method: 33\%) in~\cref{tab:model_spec}.
To this end,  we first design two compressed U-Nets, KOALA-1B and KOALA-700M, using not only block removal but also \textit{layer-wise removal} to reduce the model size of the SDXL U-Net by up to 54\% and 69\%~(vs. BK's method: 33\%) in~\cref{tab:arch}.
Then, we explore how to enhance the generative capability of the compact U-Net based on three empirical findings;
(1) \textbf{self-attention based knowledge distillation}: we investigate how to effectively distill SDXL as a teacher model and find \textit{essential factors} for feature-level KD.
Specifically, we found that self-attention features are the most crucial for distillation since self-attention-based KD allows models to learn more discriminative representations between objects or attributes.
(2) \textbf{Data}: When performing KD-training, high-resolution images and longer captions are more critical, even with fewer samples, than a larger number of low-resolution images with short captions.
(3) \textbf{Teacher model}: The teacher model with higher performance improves the capability of the student model, and step-distilled teachers allow the student model to reduce the denoising steps, which results in further speed-up.

% Experiments summary
Based on these findings, we train efficient text-to-image synthesis models on \textit{publicly available} LAION-POP~\cite{LAION_POP} by using two types of distilled teacher models, SDXL-Turbo~\cite{sauer2023sdxl-turbo} and SDXL-Lightning~\cite{lin2024lightning}, with 512px and 1024px resolutions, respectively.
We observe that our KD method consistently outperforms the BK~\cite{kim2023bksdm} method in both U-Net and Diffusion Transformer backbones in~\cref{tab:bk,tab:pixart}.
In addition, KOALA-Lightning-700M outperforms SSD-1B~\cite{gupta2024ssd}, which is trained using the BK method, at $3\times$ faster speed.
Furthermore, KOALA-Lightning-700M achieves $4\times$ faster speed and $3\times$ model efficiency than SDXL-Base while exhibiting satisfactory generation quality.
% Our efficient KOALA models consistently outperform BK-SDM~\cite{kim2023bksdm}'s KD methods.
% Furthermore, our smaller model, KOALA-700M, shows better performance than SDM-v2.0~\cite{sdm-v2.0}, which is one of the most widely used in the community, while having a similar model size and inference speed.
Lastly, to validate its practical impact, we perform inference analysis on a variety of \textbf{\textit{consumer-grade} GPUs} with different memory sizes~(\eg, 8GB, 11GB, and 24GB), and the results show that SDXL models cannot be mounted on an 8GB GPU, whereas our KOALA models can operate on it, demonstrating that our KOALA models are cost-effective alternatives for practitioners\footnote{\url{https://civitai.com/}} in resource-constrained environments.

% We can expect that by replacing SDXL's U-Net, our efficient KOALA U-Net backbones can be easily integrated into the most recent step-distillation methods~(\eg, SDXL-Turbo~\cite{sauer2023sdxl-turbo} and LCM~\cite{luo2023lcm}), which leads to further speed-up.

% Contribution Summary
Our main contributions are as follows:
\begin{enumerate}[left=-0.01cm]
    % \vspace{-0.2cm}

    \item We design two efficient denoising U-Net with model sizes~(1.13B/782M) that are more than twice as compact SDXL's U-Net~(2.56B).

    \item We perform a comprehensive analysis to build efficient T2I models, drawing three key lessons: self-attention distillation, the impact of data characteristics, and the influence of the teacher model.

    \item We conduct a systematic analysis of inference on consumer-grade GPUs, highlighting that our KOALA models operate efficiently on an 8 GB GPU, unlike other state-of-the-art models.

\end{enumerate}
% \input{sec/2_related_works}
% \input{figures/model_spec}
% \section{Prerequisite: SDXL}\label{sec:sdxl}
\section{In-depth Analysis: Stable Diffusion XL}\label{sec:sdxl}

\begin{figure*}[t!]
\centering
\includegraphics[width=\textwidth]{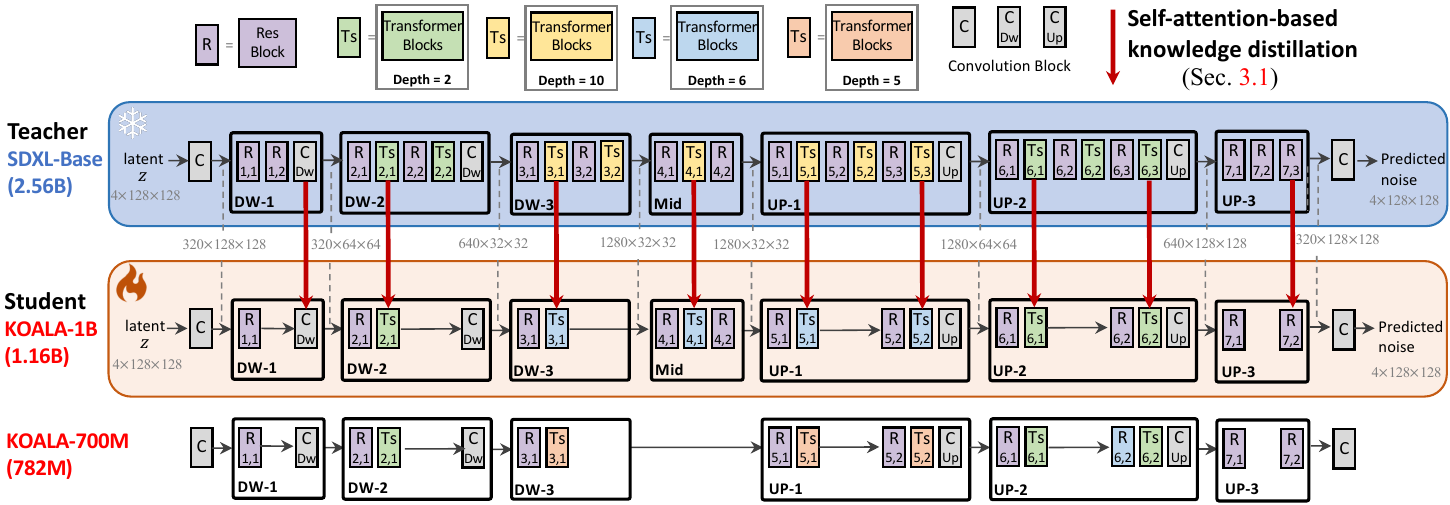}
% \vspace{0.1cm}
\caption{\textbf{Overview of \textcolor{red}{K}n\textcolor{red}{O}wledge-Distill\textcolor{red}{A}tion in \textcolor{red}{LA}tent diffusion model based on SDXL and architecture of \textcolor{red}{KOALA}.} We omit skip connections for simplicity.
% We perform feature distillation in transformer blocks using the output of the self-attention layer. 
We perform feature distillation in transformer blocks using self-attention layers. 
% (see ~\cref{subsec:kd} for details)
}
% \vspace{-0.2cm}
\label{fig:arch}
\end{figure*}

\begin{table}[ht]
    \centering
    \begin{minipage}{0.35\textwidth}
        \hspace*{-0.5cm} % Move the table to the left
        \vspace{-0.4cm}
        \caption{\textbf{SDXL-Base-1.0 model budget.} Latency is measured under the image scale of $1024^2$, FP16-precision, and 25 denoising steps in NVIDIA 4090 GPU~(24GB).}\label{tab:model_spec}
        \centering
        \vspace{-0.15cm}
        \hspace{-0.52cm}
        \scalebox{0.73}{
            \begin{tabular}{l|ccc}
            % \toprule
            SDXL      & Text Enc. & VAE Dec. & U-Net  \\ \midrule
            \#Param.  & 817M          & 83M          & 2,567M  \\
            Latency~(s) & 0.008        & 0.002       & 3.133  \\ 
            \end{tabular}
        }
    \end{minipage}
    \hfill
    \begin{minipage}{0.62\textwidth}
        \vspace{-0.1cm}
        \hspace*{-0.5cm} % Move the table to the left
        \caption{\textbf{U-Net Comparison.} Tx means Transformer. 
        % While SDM-v2.0 uses 4-stage, SDXL and KOALA~(ours) uses 3-stage removing the lowest level~($\times 8$ downsampling). 
        SDM-v2.0~\cite{sdm-v2.0} uses $768^2$ resolution, while SDXL and KOALA models use $1024^2$ resolution.
        % Latency is measured with FP16, and 25 denoising steps in NVIDIA 4090 GPU.
        CKPT means the trained checkpoint file.}\label{tab:arch}
        \vspace{-0.2cm}
        \centering
        \hspace*{-0.2cm} % Move the table to the left
        \scalebox{0.70}{
            \begin{tabular}{l|ccccc}
            U-Net & SDM-v2.0       & SDXL-1.0         &BK-SDXL & \textbf{KOALA-1B} & \textbf{KOALA-700M} \\ \midrule
            \#Param.     & 865M              & 2,567M      &1,717M    & 1,161M             & 782M                 \\
            CKPT size & 3.46GB & 10.3GB & 6.8GB & 4.4GB & 3.0GB \\
            Tx blocks & {[}1, 1, 1, 1{]} & {[}0, 2, 10{]} & {[}0, 2, 10{]} & {[}0, 2, 6{]}     & {[}0, 2, 5{]} \\
            Mid block & \checkmark & \checkmark & \checkmark & \checkmark & \ding{55} \\ 
            % Latency & 2.375s & 5.888s & 3.026s & 2.400s \\
            Latency & 1.13s & 3.13s & 2.42s & 1.60s & 1.25s \\
            \end{tabular}
        }
    \end{minipage}
\end{table}

 SDXL~\cite{podell2023sdxl}, the latest version of the SDM series~\cite{sdm-v1.4, sdm-v2.0, rombach2022ldm}, exerts a significant influence on both the academic community and the industry due to its unprecedented $1024^2$ high-resolution image quality and open source resources.
It has several key improvement points from the previous SDM-v2.0~\cite{sdm-v2.0}, \eg, multiple sizes- \& crop-conditioning, an improved VAE, a much larger U-Net, and an ad hoc style of refinement module, which leads to significantly improved generation quality.
However, the significant enlargement of U-Net in model size results in increased computational costs and significant memory (or storage) requirements, hampering the accessibility of SDXL.
% making it challenging to use SDXL easily.
Thus, we investigate the U-Net in SDXL to design a more lightweight U-Net for knowledge distillation.
We dissect the components of SDXL, quantifying its size and latency during the denoising phase, as detailed in ~\cref{tab:model_spec}.
The enlarged U-Net~(2.56B) is the primary cause of the increasing SDXL model size~(vs. SDM-v2.0~(865 M)).
Furthermore, the latency of U-Net is the main inference time bottleneck in SDXL.
Therefore, it is necessary to reduce U-Net's model budget for better efficiency.

The SDXL's U-Net varies in the number of transformer blocks for each stage, unlike SDM-v2.0, which employs a transformer block for each stage~(see~\cref{tab:arch}).
% Unlike SDM-v2.0, which has one Transformer block at each stage, SDXL's U-Net has a different number of Transformer blocks, as shown in ~\cref{tab:arch}.
At the highest feature levels~(\eg, \texttt{DW-1\&UP-3} in~\cref{fig:arch}), SDXL uses only residual blocks without transformer blocks, instead distributing more transformer blocks to lower-level features. 
So, in~\cref{fig:unet}, we analyze the parameter distribution of each stage in the U-Net.
Most parameters~(83\%) are concentrated on the transformers with ten blocks in the lowest feature map~(\eg, $32^2$ of \texttt{DW-3}, \texttt{Mid}, \texttt{UP-1} in~\cref{fig:arch}), making the main parameter bottleneck.
Thus, it is essential to address this bottleneck when designing an efficient U-Net architecture.

\section{Three lessons for building an efficient text-to-image model}\label{sec:lesson}

In this section, we introduce three empirical lessons to realize an efficient text-to-image synthesis; first, we design a lightweight U-Net architecture and perform a comprehensive analysis with the proposed efficient U-Net to find knowledge-distillation~(KD) strategies in~\cref{sec:kd}. 
Secondly, we investigate the training data characteristics that affect image generation quality in~\cref{sec:data}. 
Finally, we explore how different teacher models influence the student model in~\cref{sec:teacher}.
\noindent
For these empirical studies, we adopt two evaluation metrics, Human Preference Score~(HPSv2)~\cite{wu2023hpsv2} and CompBench~\cite{huang2023compbench}, instead of FID.
Recently, several works~\cite{betzalel2022fid_not_good,wu2023hpsv2,podell2023sdxl} have claimed that FID~\cite{heusel2017fid} is not closely correlated with visual fidelity.
HPSv2~\cite{wu2023hpsv2} is for a visual aesthetics metric, which allows us to evaluate visual quality in terms of more specific types.
As an image-text alignment metric, Compbench~\cite{huang2023compbench} is a more comprehensive benchmark for evaluating the compositional text-to-image generation capability than the single CLIP score~\cite{hessel2021clipscore}.
We report average scores for HPS and Compbench, respectively.

% \textbf{Keyword Suggestion: Four ways for effective architectural compression, 1) Efficient U-Net Architecture, 2) Self-attention based Knowledge distillation, 3) Effective Data for architectural compression, 4) Synerge with Step-distilled Teacher}

\subsection{Lesson. 1: Self-attention based Knowledge Distillation with Efficient U-Net Architecture}
\subsubsection{Efficient U-Net architecture}\label{subsec:arch}
A prior strategy for compressing the text-to-image model is to remove a pair of residual and transformer blocks at each stage, namely block removal~\cite{kim2023bksdm, gupta2024ssd}. 
While this method may be sufficient for compressing shallow U-Nets like in SDM-v1.4~\cite{sdm-v1.4}, it shows limited effectiveness for recent, more complex U-Nets. 
For cases of SDXL~\cite{podell2023sdxl}, the compression rate is reduced only from 2.5B to 1.7B, as shown in~\cref{tab:arch}. 
To address this limitation, we investigate a new approach for compressing these heavier U-Nets to achieve more efficient text-to-image generation.

\textbf{Transformer layer-wise removal is the core of efficient U-Net architecture design.}
According to the discussion in~\cref{sec:sdxl}, the majority of parameters are concentrated in the transformer blocks at the lowest feature levels. 
Each block comprises multiple consecutive transformer layers, specifically ten layers per block in SDXL (see \cref{fig:arch}). 
We address this computational bottleneck by reducing the number of transformer layers (i.e., depth), a strategy we term \textit{layer-wise removal}.

Using this removal strategy as the core, we instantiate two compressed U-Net variants: KOALA-1B and KOALA-700M.
First, we apply the prior block-removal strategy~\cite{kim2023bksdm} to the heavy U-Net of SDXL.
We note that in the decoder part~(\eg, \texttt{UP-1} to \texttt{UP-3}), we remain more blocks than in the encoder because the decoder part plays a more important role in knowledge distillation, which is addressed in~\cref{sec:kd} and~\cref{tab:distill_loc}.
On this block-removed backbone, we then adopt \textit{layer-wise removal} at different ratios. 
Specifically, we reduce the transformer layers at the lowest feature level (\ie, \texttt{DW-3}, \texttt{Mid} and \texttt{UP-1} in~\cref{fig:arch}) from 10 to 5 for KOALA-700M and to 6 for KOALA-1B. 
For KOALA-700M, we also removed the \texttt{Mid} block. 
An overview of the compressed U-Nets is presented in ~\cref{tab:arch} and~\cref{fig:arch}. 
Our KOALA-1B model has 1.16B parameters, making it half the size of SDXL (2.56B). 
KOALA-700M, with 782M parameters, is comparable in size to SDM-v2.0 (865M). 
% Notably, the inference speed of KOALA-700M is twice as fast as SDXL and comparable to SDM-v2.0, which generates lower-resolution images. 
% Additional implementation details, including weight initialization and the strategy for selecting remaining transformer layers, are provided in the supplementary materials. \todo{TODO: Link to supple}

% \input{tables/ab_neurips}

\subsubsection{Exploring Knowledge Distillation for SDXL}\label{sec:kd}

\begin{wraptable}{r}{0.45\linewidth}
% \begin{table}[t]
    \vspace{-0.6cm}
    \caption{
    \textbf{Analysis of feature level knowledge distillation of U-Net in SDXL~\cite{podell2023sdxl}.} 
    % SA, CA, and FFN denote self-attention, cross-attention, and feed-forward net in the transformer block. 
    % Res is a convolutional residual block and LF denotes the last feature~(same in BK~\cite{kim2023bksdm}).
    % For the ablation study, we train our KOALA-1B as student U-Net for 30K iterations with a batch size of 32. 
    % We use HPSv2~\cite{wu2023hpsv2} as a visual aesthetics metric, which is more correlated with human preference than FID~\cite{heusel2017fid}.
    }
    \label{tab:ablation}
    \vspace{-0.2cm}
    \centering
    \begin{tabular}{ll}
        \begin{minipage}{0.2\textwidth}
            \hspace{-0.4cm}
            \centering
            % \scriptsize
            \fontsize{8}{10}\selectfont
            \begin{tabular}{l c}
            Type & HPSv2 \\ \midrule
            Baseline & 25.53 \\ \hline
            SA	& \textbf{26.74}  \\
            CA	& 26.11  \\
            Res	& 26.27  \\
            FFN	& 26.48  \\
            LF	& 26.63  \\
            \end{tabular}
            \hspace{-0.2cm}
            \subcaption{\textbf{Distillation type}}\label{tab:distill_type}
        \end{minipage} &
        \hspace{-0.3cm}
        \begin{minipage}{0.25\textwidth}
            \centering
            \fontsize{8}{10}\selectfont
            \begin{tabular}{l c}
            Loc. & HPSv2 \\ \midrule
            Baseline & 25.53  \\ \hline
            DW-2	& 25.32  \\
            DW-3	& 25.57 \\
            Mid	& 25.66 \\
            UP-1	& \textbf{26.52} \\
            UP-2	& 26.05 \\
            \end{tabular}
            \subcaption{\textbf{Distill stage} }
            \label{tab:distill_loc}
        \end{minipage}
    \end{tabular}
    \vspace{-0.6cm}
% \end{table}
\end{wraptable}

Prior work~\cite{kim2023bksdm} that attempts to distill an early series of stable diffusion (\ie, SDM-v1.4~\cite{sdm-v1.4}) directly follows traditional knowledge distillation literature~\cite{romero2014featkd1,heo2019featkd2}.
The compressed student U-Net model $S_\theta$ is jointly trained to learn the target task and mimic the pre-trained U-Net of SDM-v1.4 as a teacher network.
Here, the target task is the reverse denoising process~\cite{ho2020ddpm}, and we denote the corresponding learning signal as $\loss{task}$.
Besides the task loss, the compressed student model is trained to match the output of the pre-trained U-Net at both output and feature levels.
$\loss{out}$ and $\loss{feat}$ represent the knowledge distillation (KD) loss at the output- and feature-level, respectively.
% The knowledge distillation (KD) loss at output- and feature-level, $\loss{out}$ and $\loss{feat}$, are formulated as:
% \begin{equation}
%     \loss{outKD} = \min_{S_\theta} \expect_{z, \epsilon, t, c}||\epsilon_{T_\theta}(z, t, c) - \epsilon_{S_\theta}(z, t, c)||^2_2,
% \end{equation}\label{eq:out_loss}
% \begin{equation}
%     \loss{featKD} = \min_{S_\theta} \expect_{z, \epsilon, c, t}||\sum_i{f_T^i(z_t, t, c)-f_S^i(z_t, t, c)}||^2_2,
% \end{equation}\label{eq:feat_loss}
% where $\epsilon_{T_\theta}(\cdot)$ denotes the predicted noise from each U-Net in the teacher model.
For designing the feature-level KD-loss, BK-SDM~\cite{kim2023bksdm} simply considers only the last feature (\texttt{LF}) map of the teacher $f_T^i(\cdot)$ and student network $f_S^i(\cdot)$ at each stage as follows:
\begin{equation}
    \loss{featKD} = \min_{S_\theta} \expect_{z, \epsilon, c, t}||\sum_i{f_T^i(z_t, t, c)-f_S^i(z_t, t, c)}||^2_2,
\end{equation}\label{eq:feat_loss}
% \vspace{-0.1cm}

\noindent
where $t$ and $c$ denote given diffusion timestep and text embeddings as conditions.
% $\epsilon_{T_\theta}(\cdot)$ denotes the predicted noise from each U-Net in the teacher model.
Thus, the feature distillation approach for text-to-image diffusion models has \textbf{\textit{not been sufficiently explored}}, leaving room for further investigation.

% In this work, we extensively explore feature distillation strategies to distill the knowledge from the U-Net of SDXL effectively to our efficient U-Net, KOALA-1B.
In this section, we extensively explore feature distillation strategies to distill the knowledge from the U-Net of SDXL effectively to our efficient U-Net, KOALA-1B.
We start from a baseline trained only by $\loss{task}$ and add $\loss{featKD}$ without $\loss{outKD}$ to validate the effect of feature distillation.
More training details are described in~\cref{subsec:ex_details} and \cref{sec:app_impl}.
% From the experiments as shown in~\cref{tab:ablation}, We summarize our insights into \textit{\textbf{four important findings}} as follows.

\begin{figure*}[t]
\centering
\includegraphics[width=\textwidth]{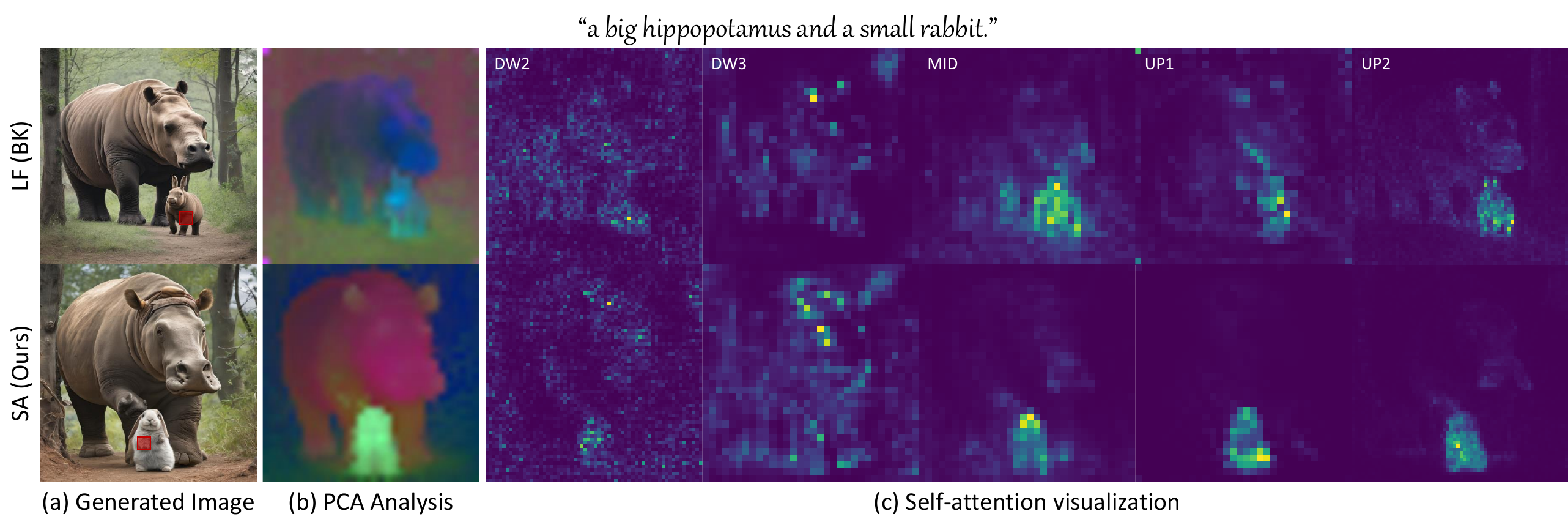}
% \vspace{-0.5cm}
\caption{ \textbf{Analysis on self-attention maps of distilled student U-Nets.} (a) Generated images of LF- and SA-based distilled models, which are BK-SDM~\cite{kim2023bksdm} and our proposal, respectively. In BK-SDM's result, a rabbit is depicted like a hippopotamus (\ie, appearance leakage). (b) Visualization of PCA analysis results on self-attention maps of \texttt{UP-1} stage. (c) Representative visualization of self-attention map from different U-Net stages. Red boxes denote the query patches.
Note that from the \texttt{MID} stage, the \texttt{SA}-based model \textbf{\textit{attends}} to the rabbit more \textit{\textbf{discriminatively}} than the \texttt{LF} model, demonstrating that self-attention-based KD allows to generate objects more distinctly. 
% More visualizations are in the supplementary material.
}
% \vspace{-0.5cm}
\label{fig:attention_analysis}
\end{figure*}
%Generated images of LF- and SA-based distilled models, which are BK-SDM's and our proposal, respectively.

\noindent
% \textbf{F1. Which feature type is effective for distillation?}
\textbf{Self-attention-based knowledge distillation transfers discriminative image representation.}

With the increasing complexity of U-Net and its stage, relying solely on the last feature~(\texttt{LF}) as in BK~\cite{kim2023bksdm} may not be sufficient to mimic the intricate behavior of the teacher U-Net.
Thus, we revisit which features provide the richest guidance for effective knowledge distillation.
We focus on key intermediate features from each stage: outputs from the
self-attention~(\texttt{SA}), cross-attention~(\texttt{CA}), and feedforward net~(\texttt{FFN}) in the transformer block, as well as outputs from convolutional residual block~(\texttt{Res}) and \texttt{LF}.
~\cref{tab:distill_type} summarizes the experimental results.
While all types of features help obtain higher performance over the naïve baseline with only the task loss, distilling \textbf{\textit{self-attention features}} achieves the most performance gain.
Considering the prior studies~\cite{kolkin2019style,shechtman2007matching,tumanyan2022splicing} which suggest that \texttt{SA} plays a vital role in capturing semantic affinities and the overall structure of images, the results emphasize that such information is crucial for the distillation process. \\
\indent To understand the effects more clearly, we illustrate a representative example in the~\cref{fig:attention_analysis}.
To reason about how the distilled student U-Net captures self-similarity, we perform a PCA analysis~\cite{jolliffe2016principal,tumanyan2023plug} on self-attention maps.
Specifically, we apply PCA on self-attention maps from \texttt{SA}- and \texttt{LF}-based models and show the top three principal components in~\cref{fig:attention_analysis}-(b).
Interestingly, in the \texttt{SA}-based model, each principal component distinctly represents individual objects (\ie, unique color assignments to each object).
This indicates that the \texttt{SA}-based model effectively distinguishes different objects in modeling self-similarity, which plays a crucial role in accurately rendering the distinct appearance of each object.
In contrast, the \texttt{LF}-based model exhibits less distinction between objects, resulting in \textit{appearance leakage} between them~(\eg, a small hippo with rabbit ears). More PCA analyses are detailed in~\cref{fig:supple_attn_viz}.

\noindent
% \textbf{F2. Which stage is most effective for distillation?}
\textbf{Self-attention at the decoder has a larger impact on the quality of generated images.}

We further explore the role and significance of each self-attention stage. 
To this end, we first visualize the self-attention map in ~\cref{fig:attention_analysis}-(c).
The self-attention maps initially capture general contextual information (\eg, \texttt{DW-2\&DW-3}) and gradually focus on localized semantics (\eg, \texttt{MID}). 
In the decoder, self-attentions increasingly correlate with higher-level semantics (\eg, object) to accurately model appearances and structures.
Notably, in this stage, the \texttt{SA-}based model attends corresponding object regions~(given the query patch, red box) more \textit{discriminatively} than the \texttt{LF-}based model, which results in improved compositional image generation performance.

% Building on this insight, 
In addition, we ablate the significance of each self-attention stage in the distillation process.
Specifically, we adopt an \texttt{SA-}based loss at a single stage alongside the task loss.
As shown in \cref{tab:distill_loc}, the results align with the above understanding: distilling self-attention knowledge within the \textbf{\textit{decoder}} stages significantly enhances generation quality. 
In comparison, the impact of self-attention solely within the encoder stages is less pronounced. 
Consequently, we opt to retain more \texttt{SA} layers within the decoder (see \cref{fig:arch}).

In summary, we train our efficient KOALA U-Nets using the following objectives: $\loss{task} + \loss{outKD} + \loss{featKD}$. 
We apply our proposed self-attention-based knowledge distillation (KD) methods to $\loss{featKD}$. 
% Additional implementation details and analysis, including the synergy between the proposed self-attention-based KD and traditional KD, are provided in the supplementary. \todo{TODO: Link to supple}
Further analyses on featKD, including how to locate features and combine different types of features, are provided in~\cref{sec:app_kd}.

% We train our KOALA models with the following objectives: $\loss{task} + \loss{outKD} + \loss{featKD}$ where we apply our findings to $\loss{featKD}$. 
% As shown in~\cref{tab:arch}, we design two models, KOALA-1B and KOALA-700M, based on SDXL~\cite{podell2023sdxl}, with U-Net model sizes of 1.16B and 782M, respectively.

\subsection{Lesson 2. Data: the impact of sample size, image resolution, and caption length}\label{sec:data}

\begin{table}[t]
    \centering
    \hspace*{-0.2cm} % Move the table to the left
    \begin{minipage}[t]{0.63\textwidth}
        \centering
        \vspace{-0.3cm}
        \caption{\textbf{Training Data} comparison. 
        AR and ACL mean average resolution and average caption length, respectively. \texttt{synCap} means synthetic captions by LLaVA-v1.5~\cite{liu2023improvedllava}}\label{tab:data}
        \vspace{+0.1cm}
        \hspace*{-0.3cm} % Move the table to the left
        \scalebox{0.72}{
            \begin{tabular}{llccccc}
            % Data                 & \#Samples & avg. Resolution & avg.Prompt Length & HPSv2 & CompBench \\ \midrule
            ID & Data                 & \#Imgs & AR & ACL & HPSv2 & CompBench \\ \midrule
            (a) & LAION-A-6+~\cite{laion-aesthetics-6plus}  & 8M  & $580\times676$ & 13 & 27.43 & 0.3791 \\
            % (b) LAION-A-6+ w/ \texttt{synCap} & 8M                   & $580\times676$                             & 72                                 & 27.61                     & 0.4168                        \\
            (b) & (a) + \texttt{synCap} & 8M                   & $580\times676$                             & 72                                 & 27.61                     & 0.4168                        \\
            (c) & LAION-POP~\cite{LAION_POP}            & 491K                     & $1274\times1457$                           & 81                                 & \textbf{27.79}            & \textbf{0.4290}               \\
            \\
            \\
            \end{tabular}
        }
    \end{minipage}
    \vspace{-0.4cm}
    \hfill
    \begin{minipage}[t]{0.36\textwidth}
        \hspace*{-0.3cm} % Move the table to the left
        \caption{\textbf{Teacher model} comparison. We use KOALA-700M as a student model.
        }\label{tab:teacher}
        \vspace{-0.25cm}
        \centering
        \hspace*{-0.3cm} % Move the table to the left
        \scalebox{0.73}{
            \begin{tabular}{lccc}
            Teacher model & Step & HPSv2 & CompBench \\ \midrule
            SDXL-Base-1.0	& 25 & 27.79 & 0.4290 \\
            SDXL-Turbo	& 10 & 27.88	&0.4470 \\
            SDXL-Lightning	 & 10 & \textbf{28.13} & \textbf{0.4538} \\
            \end{tabular}
        }
    \end{minipage}
    \vspace{-0.4cm}
\end{table}
% \vspace{-0.3cm}

We investigate various data factors—such as image resolution, caption length, and the number of samples—that impact the quality of the final text-to-image model. 
To ensure reproducibility, we design three data variants using open-source data. 
(i) LAION-Aesthetic-6+~(LAION-A-6+)~\cite{laion-aesthetics-6plus} includes a large volume of image-text pairs (8,483,623) with images filtered for high aesthetics. 
Most images are low-resolution (average $580\times676$), and the corresponding captions are brief (average length of 13 words).
(ii) Description-augmented LAION-A-6+ is designed to demonstrate the impact of detailed descriptions. 
For each image in LAION-A-6+, we use a large multimodal model~\cite{liu2023improvedllava} (LMM) to generate detailed descriptions. 
These synthesized captions, referred to as \texttt{synCap}, convey significantly more semantic information and are longer (e.g., an average length of 72 words; more details on \texttt{synCap} in~\cref{sec:app_llava}). This data source is denoted in the second row of the table.
(iii) LAION-POP~\cite{LAION_POP} features high-resolution images (average $1274\times1457$) and descriptive captions (average length of 81 words), although the dataset size is relatively small (e.g., 491,567 samples). 
The descriptions are generated by LMM, CogVML~\cite{wang2023cogvlm} and LLaVA-v1.5.

We train KOALA-700M models using the same training recipes for each data source. 
From the results summarized in~\cref{tab:data}, we make several observations. 
First, detailed captions significantly boost performance, enabling the model to learn detailed correspondences between images and text (See (a) and (b) in the Compbench score).
Second, high-resolution images, which convey complex image structures, are a valuable source for training T2I models. 
Despite having fewer samples, LAION-POP further boosts overall performance. 
Based on these findings, we opt to use LAION-POP as the main training data, as it features high-resolution images and descriptive captions.

\subsection{Lesson 3. The influence of Teacher model}\label{sec:teacher}
Following the tremendous success of SDXL~\cite{podell2023sdxl}, recent large-scale text-to-image models have adopted its U-Net backbone. 
SDXL-Turbo~\cite{sauer2023sdxl-turbo} and SDXL-Lightning~\cite{lin2024lightning} are notable examples, enabling high-quality image generation in low-step regime through progressive distillation~\cite{salimans2022progressive}. 
This section investigates whether our distillation framework can effectively exploit these diverse models. 
To this end, we leverage SDXL-Base and its variants, SDXL-Turbo and SDXL-Lightning, as teacher models, transferring their knowledge into KOALA-700M.
We apply the former two lessons and more training details are described in~\cref{sec:app_training}.

\begin{wrapfigure}{r}{0.4\textwidth}
    \centering
    % \vspace{-0.2cm}
    \includegraphics[width=\linewidth]{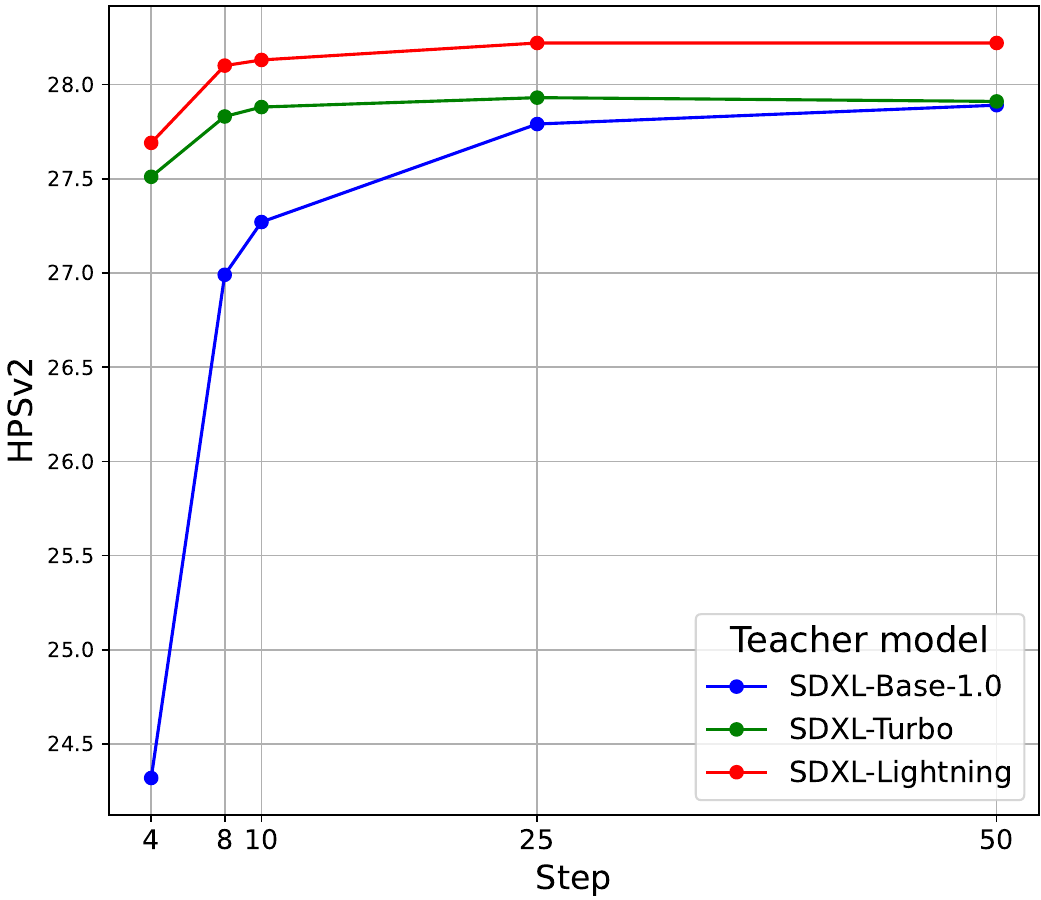}
    \vspace{-0.6cm}
    \caption{{Teacher model comparisons across denoising steps with KOALA-700M.}} \label{fig:step}
    % \vspace{-0.55cm}
\end{wrapfigure}
\begin{figure*}[t]
    \begin{minipage}[t]{.53\textwidth}
        \centering
        \includegraphics[width=\textwidth]{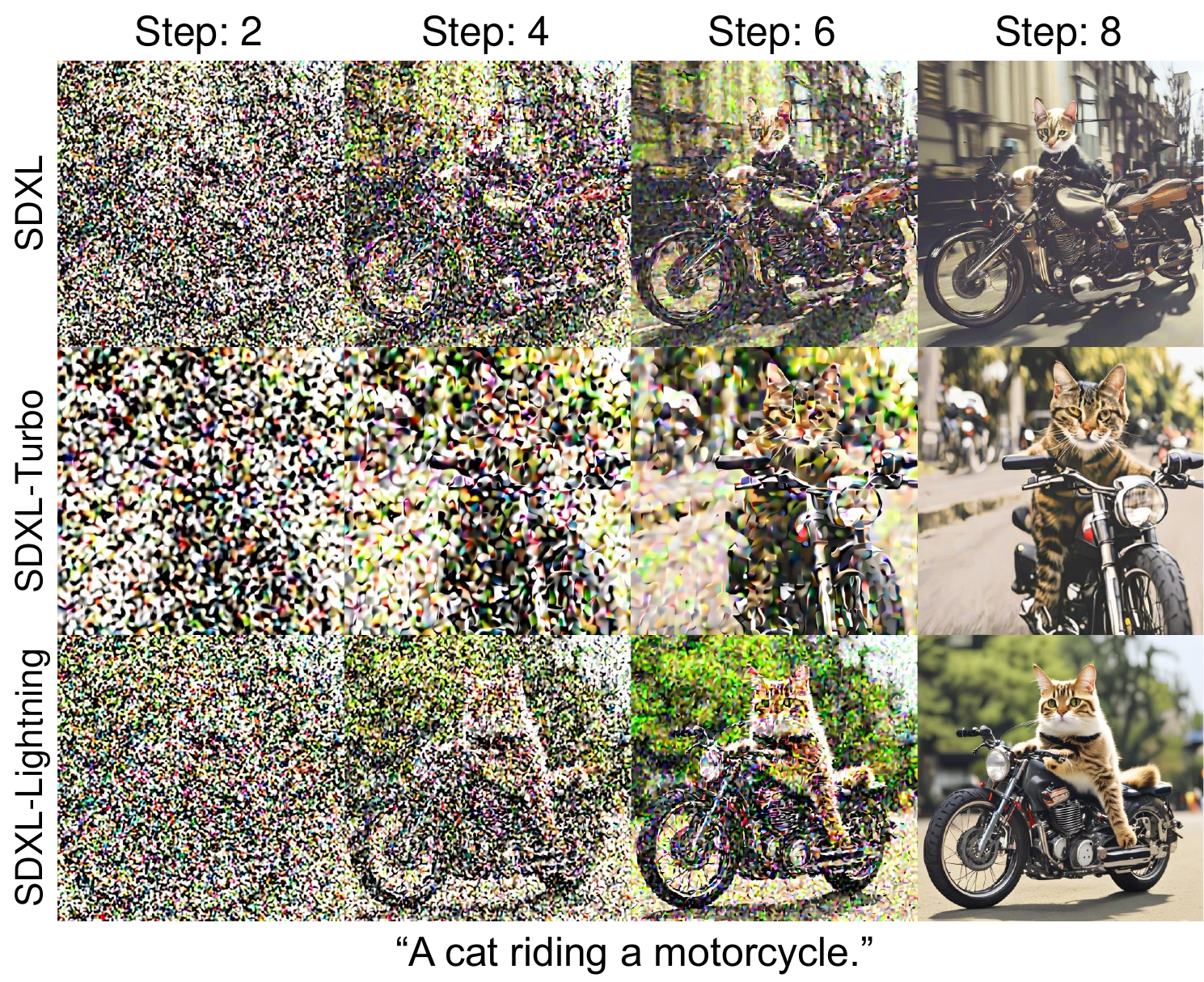}
        \caption{\footnotesize Denoising process of different teacher models.}\label{fig:1}
        \label{fig:teacher_abl}
    \end{minipage}
    \hfill
    \begin{minipage}[t]{.46\textwidth}
        \centering
        \includegraphics[width=\textwidth]{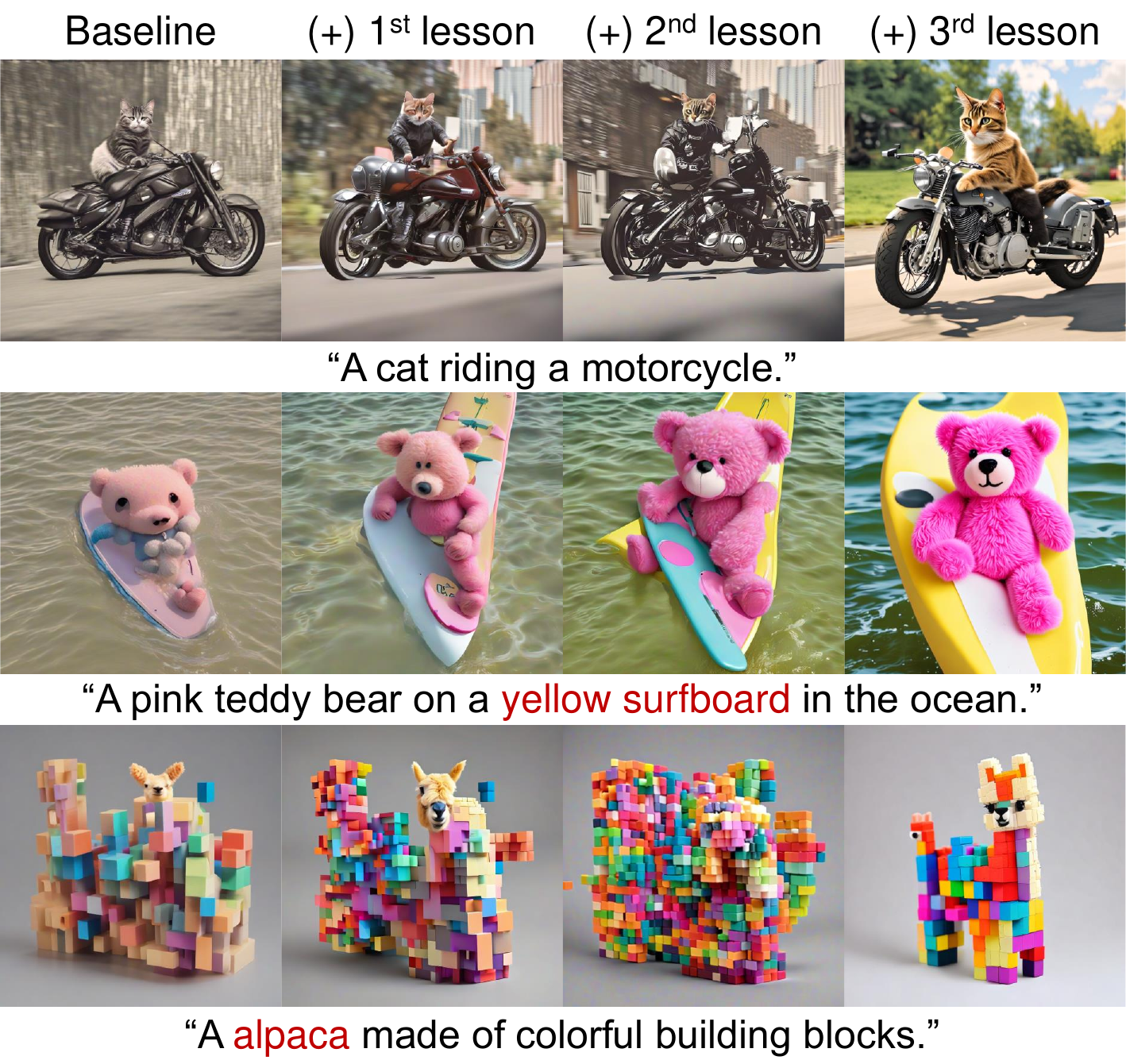}
        \caption{\footnotesize Qualitative analysis on proposed lessons.}\label{fig:2}
        \label{fig:lesson_abl}
    \end{minipage} \label{fig:1-2}
    % \caption{Title.}
    \vspace{-0.2cm}
\end{figure*}

As shown in \cref{tab:teacher}, all KOALA-700M models distilled from different teacher models demonstrate decent image generation capabilities. This highlights the generality of our knowledge distillation framework. 
More interestingly, when using SDXL-Turbo and SDXL-Lightning as teachers, KOALA-700M models exhibit comparable or even better image quality than when SDXL-Base is used as the teacher, despite requiring fewer denoising steps (\eg, 10 vs. 25).
Note that the KD framework or noise schedule (Euler discrete schedule~\cite{karras2022elucidating}, the same as SDXL-Base) for the different KOALA models does not require specific modifications.
Thus, KOALA models seamlessly inherit the ability to illustrate realistic structures and details in images, even in the short-step regime, from their step-distilled teachers (See 2$^{nd}$ and 3$^{rd}$ rows in \cref{fig:teacher_abl}). 
This results in robust performance across a diverse range of denoising steps (See \cref{fig:step}). 
In contrast, when SDXL is used as the teacher model, KOALA models struggle to depict realistic structures in a few steps, leading to flawed features (e.g., missing handles and lights on a motorcycle in \cref{fig:teacher_abl}).
For more efficient text-to-image synthesis, we leverage step-distilled teachers, enabling KOALA models to generate high-quality images in just a few steps.

Therefore, combining all these lessons, we build \textbf{K}n\textbf{O}wledge-distill\textbf{A}tion-based \textbf{LA}tent diffusion models, called KOALA. 
As discussed in each section, the three proposed lessons complement each other. 
Both image quality and image-text alignment improve progressively as each lesson is added, as shown in \cref{fig:lesson_abl}.
In particular, we build two types of KOALA models: KOALA-Turbo with 512px and KOALA-Lightning with 1024px using two KOALA U-Net~(1B\&700M), respectively.
% with our efficient U-Nets

\section{Experiments}
% \vspace{-0.2cm}

\subsection{Implementation details} \label{subsec:ex_details}

\noindent
\textbf{Dataset.}
As the datasets used for state-of-the-art methods in~\cref{tab:main} are proprietary or not released, we opt for the \textit{publicly accessible} only LAION dataset~\cite{schuhmann2022laion-5b} to ensure the reproducibility of our work.
From the data recipe in~\cref{sec:data}, we finally use LAION-POP~\cite{LAION_POP} for KOALA models in~\cref{tab:main}.
More details of the dataset we used are described in~\cref{sec:app_data}

\noindent
\textbf{Training.}
% We base on the official SDXL-Base-1.0~\cite{huggingface_sdxl} model and their training settings while only replacing its U-Net with our efficient U-Net.
According to the recipe about the Teacher model in~\cref{sec:teacher}, we use SDXL-Turbo~\cite{sauer2023sdxl-turbo} and SDXL-Lightning~\cite{lin2024lightning} as teacher models, building two types of KOALA models, KOALA-Turbo and KOALA-Lightning.
Since SDXL-Turbo and -Lightning are based on SDXL-Base model~\cite{podell2023sdxl}, we use the same two text encoders, OpenCLIP ViT-bigG~\cite{ilharco_openclip} and CLIP ViT-L~\cite{radford2021clip} and only replace the original U-Net with our efficient KOALA U-Net.
% For VAE, we use \texttt{sdxl-vae-fp16-fix}~\cite{vae-fp16-fix}, which enables us to use FP16 precision for VAE computation.
Our U-Nets are initialized with the teacher's U-Net weights at the exact block location.
% We train our KOALA models for 100K iterations using four NVIDIA A100~(80GB) GPUs with Euler discrete scheduler~\cite{karras2022elucidating}, a batch size of 128, AdamW optimizer~\cite{loshchilov2017adamw}, a constant learning rate of $10^{-5}$, and FP16 precision.
Using our self-attention-based KD method in~\cref{sec:kd}, we train our KOALA models on the LAION-POP dataset using four NVIDIA A100~(80GB) GPUs with $512^2$ and $1024^2$ resolutions for KOALA-Turbo and KOALA-Lightning, respectively.
\textbf{Inference.} We use Euler discrete scheduler~\cite{karras2022elucidating} as the same sampler in SDXL~\cite{podell2023sdxl}. All KOALA models generate images with 10 denoising steps, FP16, and cfg-sale~\cite{ho2022cfg} of 3.5.
Please see further details of training and inference in~\cref{sec:app_training} and~\cref{sec:app_infernce}, respectively.
% More training and inference details are described in~\cref{sec:app_impl}.

% \noindent
% \textbf{Evaluation metric.}
% Recently, several works~\cite{betzalel2022fid_not_good,wu2023hpsv2,podell2023sdxl} have claimed that FID~\cite{heusel2017fid} is not closely correlated with visual fidelity because a feature extractor for FID is pre-trained on the ImageNet dataset, which does not overlap much with the datasets used to train recent text-to-image models~(\eg, style, types, resolution, etc.).
% Therefore, instead of FID, we use \textbf{Human Preference Score~(HPSv2)}~\cite{wu2023hpsv2} as a visual aesthetics metric, which allows us to evaluate visual quality in terms of more specific types.
% For image-text alignment, we use the \textbf{Compbench}~\cite{huang2023compbench}, which is a more comprehensive benchmark for evaluating the compositional text-to-image generation capability than the single CLIP score~\cite{hessel2021clipscore}.

% \input{tables/bk_comparison}
\begin{table*}[t]
% \caption{\footnotesize \textbf{Performance comparison to state-of-the-art models}. 
\caption{\textbf{Performance comparison to state-of-the-art models}. 
We measure latency and memory usage with a bath size of 1 on NVIDIA 4090 GPU. We obtain HPSv2 and Compbench scores of all models on the same GPU and library environment by using their official weights. We highlight the \colorbox{teagreen}{best value} in green, and the \colorbox{palecornflowerblue}{second-best} value in blue. The full scores of HPSv2 and Compbench are shown in~\cref{tab:main_table_full}.}\label{tab:main}
\begin{adjustbox}{width=0.98\textwidth, center}
    % \footnotesize % 폰트 크기를 작게 조정
    \begin{tabular}{lcccrrcc}
    Model                      & Resolution & Steps & Latency~(s) & U-Net Param. & Memory & HPSv2            & CompBench       \\ \midrule
    SDM-v2.0~\cite{sdm-v2.0} &$768^2$ &25 &1.236 &0.86B &5.6GB &25.86 &0.3672 \\
    SDXL-Base-1.0~\cite{podell2023sdxl}  & $1024^2$            & 25    & 3.229   & 2.56B   & 11.9GB        & 30.82            & 0.4445              \\
    SDXL-Turbo~\cite{sauer2023sdxl-turbo}  & $512^2$               & 8     & 0.245   & 2.56B   & 8.5GB         & 29.93 & 0.4489 \\
    SDXL-Lightning~\cite{lin2024lightning}      & $1024^2$       & 8     & 0.719   & 2.56B   & 11.7GB        & \cellcolor{teagreen}32.18   & 0.4445 \\
    Pixart-$\alpha$~\cite{chen2023pixartalpha}   & $1024^2$            & 25    & 3.722   & \cellcolor{teagreen}0.6B    & 17.3GB        & 32.06            & 0.3880          \\
    Pixart-$\Sigma$~\cite{chen2024pixartsigma}   & $1024^2$            & 25    & 3.976   & \cellcolor{teagreen}0.6B    & 17.3GB        & 31.75            & \cellcolor{teagreen}0.4612          \\
    SSD-1B~\cite{gupta2024ssd}         & $1024^2$            & 25    & 2.094   & 1.3B    & 9.4GB         & 31.43            & 0.4497          \\
    SSD-Vega~\cite{gupta2024ssd}       & $1024^2$            & 25    & 1.490   & \cellcolor{palecornflowerblue}0.74B   & 8.2GB         & \cellcolor{palecornflowerblue}32.17            & 0.4461          \\ \midrule
    \textbf{KOALA-Turbo-700M}    & $512^2$       & 10    & \cellcolor{teagreen}0.194   & 0.78B   & \cellcolor{teagreen}4.9GB         & 29.98            & 0.4555          \\
    \textbf{KOALA-Turbo-1B}     & $512^2$        & 10    & \cellcolor{palecornflowerblue}0.238   & 1.16B    & \cellcolor{palecornflowerblue}5.7GB         & 29.84            & 0.4560          \\
    \textbf{KOALA-Lightning-700M}    & $1024^2$   & 10    & 0.655   & 0.78B   & 8.3GB         & 31.50            & 0.4505          \\
    \textbf{KOALA-Lightning-1B}    & $1024^2$     & 10    & 0.790   & 1.16B   & 9.1GB         & 31.71            & \cellcolor{palecornflowerblue}0.4590
    \end{tabular}
\end{adjustbox}
% \vspace{+0.1cm}
\end{table*}

\subsection{Main results}

\noindent
% \textbf{Comparison with SDXL models.}
\textbf{vs. SDXL models:}
% Compared to SDXL models, our KOALA-700M/-1B models have much smaller model sizes and require much less memory usage.
Compared to SDXL-Base-1.0~\cite{podell2023sdxl}, our KOALA-Lightning-700M/-1B models achieve better performance in terms of HPSv2 and CompBench while showing about 5$\times$ and 4$\times$ faster speed, respectively.
% Compared to SDXL-Turbo, our KOALA-Turbo models show comparable visual quality but achieve higher CompBench scores.
% Compared to SDXL-Lightning, our KOALA Lightning models show inferior HPSv2 but  better CompBench scores
% Compared to SDXL-Turbo~\cite{sauer2023sdxl-turbo} and SDXL-Lightning~\cite{lin2024lightning}, our KOALA-Turbo and KOALA-Lightning models show comparable or inferior HPSv2 but achieve higher CompBench scores with much smaller model size~(up to $3\times$) and less memory usage~(up to 1.7$\times$).
Compared to SDXL-Turbo~\cite{sauer2023sdxl-turbo} and SDXL-Lightning~\cite{lin2024lightning}, our KOALA-Turbo and KOALA-Lightning models show comparable or inferior HPSv2 scores but achieve higher CompBench scores with up to $3\times$ smaller model sizes and $1.7\times$ lower memory usage.
\noindent
% \textbf{Comparison with SSD.}
\textbf{vs. Pixart:}
% Compared to Pixart-$\alpha$~\cite{chen2023pixartalpha}\&-$\Sigma$~\cite{chen2024pixartsigma}, 
KOALA-Lightning models fall short in HPSv2 and CompBench. Especially, Pixart-$\Sigma$~\cite{chen2024pixartsigma} achieves the best CompBench and the second-best HPSv2 scores.
This result is attributed to data quality, as Pixart-$\Sigma$ collects high-quality internal data consisting of 33 million images above 1K resolution and uses synthetic longer captions (with an average length of 184 words).
However, our KOALA-Lightning-700M shows $6\times$ faster speed and $2\times$ better memory efficiency.
\textbf{vs. SSD:}
% Although SSD~\cite{gupta2024ssd} shows similar architectures with ours, KD-training method, the teacher model, and training dataset used are different and, in turn
Note that due to the difference in training datasets, we cannot make a direct comparison with SSD models, which are trained by BK~\cite{kim2023bksdm}'s KD method.
Except for HPSv2 of SSD-Vega~\cite{gupta2024ssd}, KOALA-Lighting models show better HPSv2 and CompBench scores while achieving up to $3\times$ faster speed.
More qualitative comparisons in~\cref{sec:app_samples} support the quantitative results.

\begin{table}[t]
    \centering
    \begin{minipage}{0.55\textwidth}
        \hspace*{-0.5cm} % Move the table to the left
        % \vspace{-0.4cm}
        \caption{\textbf{Comparison to BK~\cite{kim2023bksdm}}. All models are trained for 50K iterations same as BK-SDM.}\label{tab:bk}
        \centering
        \vspace{-0.2cm}
        \hspace{-0.48cm}
        \scalebox{0.85}{
            \begin{tabular}{@{}llcc@{}}
            KD method & Backbone   & HPSv2  & CompBench \\ \midrule
            BK~\cite{kim2023bksdm}        & BK-Small   & 26.72 & 0.3237    \\
            \textbf{Ours}      & BK-Small   & \textbf{26.86} & \textbf{0.3417 }   \\
            BK~\cite{kim2023bksdm}        & KOALA-1B   & 27.01  & 0.3599    \\
            \textbf{Ours}      & KOALA-1B   & \textbf{27.15}  & \textbf{0.3712 }  
            \end{tabular}
        }
    \end{minipage}
    \hfill
    \begin{minipage}{0.4\textwidth}
        % \vspace{-0.4cm}
        \hspace*{-0.7cm} % Move the table to the left
        \caption{\textbf{KD feature types in Diffusion Transformer~(Pixart-$\Sigma$~\cite{chen2024pixartsigma}}).}\label{tab:pixart}
        \vspace{-0.2cm}
        \centering
        \hspace*{-0.2cm} % Move the table to the left
        \scalebox{0.85}{
            \begin{tabular}{lcc}
            KD Type. & HPSv2 & CompBench \\\midrule
            SA & \textbf{25.16} & \textbf{0.4281} \\
            CA & 24.94 & 0.4279 \\
            FFN & 24.80 & 0.4191 \\
            LF in BK~\cite{kim2023bksdm} & 21.62 & 0.3527 \\
            \end{tabular}
        }
    \end{minipage}
    % \vspace{-0.5cm}
\end{table}

\subsection{Discussion}

\noindent
\textbf{Comparison with BK-SDM.}
For a fair comparison to BK-SDM~\cite{kim2023bksdm}, we train our KOALA U-Net backbones with their distillation method under the same data setup~(See training details in~\ref{sec:app_training}).
As shown in~\cref{tab:bk}, our KD method consistently achieves higher HPSv2 and CompBench scores than the BK-SDM~\cite{kim2023bksdm} when using different U-Net backbones.
These results demonstrate two main implications as follows:
1) the proposed distillation of the self-attention layer is more helpful for visual aesthetics than simply distilling the last layer feature by BK~\cite{kim2023bksdm}.
2) our self-attention-based KD approach allows the model to learn more discriminative representations between objects or attributes so that it can follow prompts faithfully (as shown in~\cref{sec:kd} and~\cref{fig:attention_analysis}).
More qualitative comparisons are demonstrated in~\cref{fig:bk_comparison}.

\noindent
\textbf{Applicability of self-attention based KD to Diffusion Transformer.}
To validate the generality of our self-attention distillation method, we also apply it to a diffusion transformer~(DiT) based T2I model, Pixart-$\Sigma$~\cite{chen2024pixartsigma}.
To this end, We compress the DiT-XL~\cite{peebles2023dit} backbone and build DiT-M by reducing the number of layers from 28 to 14 with the same hidden dimension~(see more training details in~\cref{sec:app_pixart}.).
Following our KD strategy, we conduct an ablation study by simply changing the distillation location due to DiT's architectural simplicity, which consists of only transformer blocks without resolution changes.
\cref{tab:pixart} illustrates that distilling the self-attention feature outperforms other features while using the last features proposed in BK~\cite{kim2023bksdm} shows the worst performance, demonstrating that the self-attention layer is still the most crucial part for diffusion transformer.

\noindent
\textbf{Synergy Effect with Step-Distillation Method, PCM~\cite{wang2024pcm}.}
Since step-distillation methods~\cite{lin2024lightning,sauer2023sdxl-turbo,wang2024pcm} and our KD approach are orthogonal, applying our KOALA backbones to the step-distillation methods could yield synergistic effects, leading to further speed improvements.
% We verify the synergy effect between the step-distillation method (\eg, PCM~\cite{wang2024pcm}) and our KOALA backbones.
To verify the synergy effect between the step-distillation method (\eg, PCM~\cite{wang2024pcm}) and our KOALA backbones, we conduct step-distillation training using PCM with our KOALA backbones, and the results are presented in~\cref{tab:pcm}.
Thanks to their efficiency, our KOALA backbones allow PCM\footnote{PCM is the only work that officially provides step-distillation training code}\textsuperscript{,}\footnote{\url{https://github.com/G-U-N/Phased-Consistency-Model}} to achieve additional speed-up with only a slight performance drop compared to using the SDXL-Base backbone.
Furthermore, we provide qualitative comparisons between PCM-KOALA models and PCM-SDXL-Base in~\cref{fig:pcm_comparison}, demonstrating that the generated images achieve visual quality comparable to those of SDXL-Base.

\begin{figure*}[t]
\centering
\includegraphics[width=0.97\textwidth]{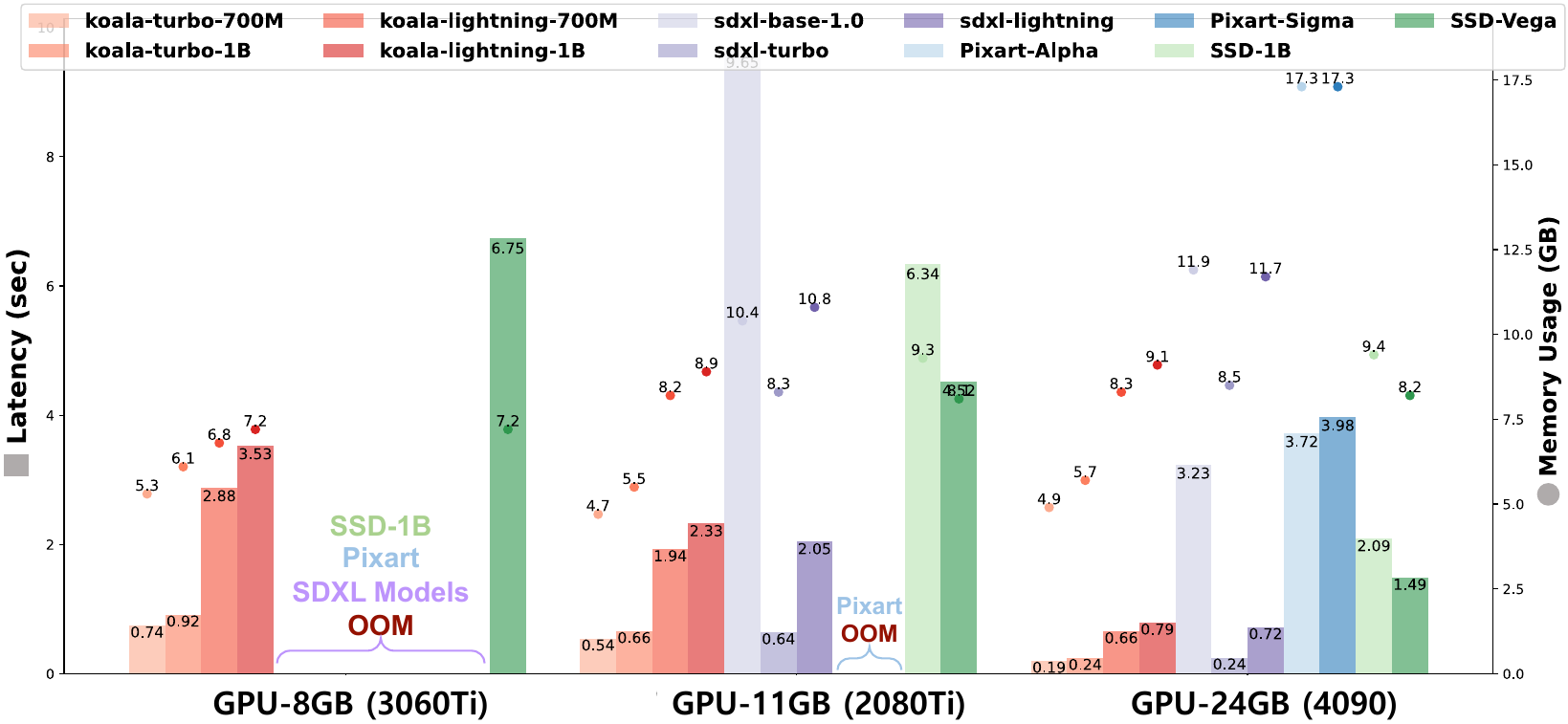}
% \vspace{-0.7cm}
\caption{\footnotesize \textbf{Latency and Memory} comparison on across different \textbf{\textit{consumer-grade} GPUs}. 
% SDXL-Turbo and KOALA-Turbo models generate images with $512^2$ resolution, while the others use $1024^2$. OOM means Out-of-Memory.
We run each model with the denoising steps in~\cref{tab:main} and FP16.
For a fair comparison, we use the official pre-trained weights and inference code in the Hugginface without any other tricks such as \texttt{torch.compile} or quantization.
% We use FP16 precision.
% For fair comparison, we measure inference and memory usage on the same GPU machine with identical 
Note that \textbf{only our KOALA models and SSD-Vega can run all types of GPUs}.
}
% \vspace{+0.1cm}
% \vspace{-0.3cm}
\label{fig:latency_memory}
\end{figure*}
\begin{table}[t]
\centering
\caption{\textbf{Synergy effect with a step-distilled method}, PCM~\cite{wang2024pcm}.
We conduct step-distillation training using PCM with our KOALA backbones and compare with PCM-SDXL-Base.
}
\vspace{+0.2cm}
\begin{adjustbox}{width=1.\textwidth, center}
\begin{tabular}{l l l c c c c c c}
Method & Teacher & Student & \#Step & Param. (B) & Memory (GB) & Latency & HPS & CompBench \\
\midrule
PCM~\cite{wang2024pcm} & SDXL-Base & SDXL-Base & 2 & 2.56 & 12.1 & 0.345 & \textbf{29.99} & \textbf{0.4169} \\
PCM~\cite{wang2024pcm} & SDXL-Base & \textbf{KOALA-700M} & 2 & \textbf{0.78} & \textbf{8.2} & \textbf{0.222} & 28.78 & 0.3930 \\
PCM~\cite{wang2024pcm} & SDXL-Base & \textbf{KOALA-1B} & 2 & 1.16 & 9.0 & 0.235 & 29.04 & 0.4055 \\
\end{tabular}
\end{adjustbox}
\label{tab:pcm}
\end{table}

\subsection{Model budget comparison on consumer-grade GPUs}\label{sec:gpus}
% \vspace{-0.15cm}

% We further validate the efficiency of our model by measuring its inference speed and memory usage on a variety of \textit{consumer-grade} GPUs with different memory sizes, such as 8GB~(3060Ti), 11GB~(2080Ti), and 24GB~(4090), because the GPU environment varies for each user.
% ~\cref{fig:latency_gpu} illustrates inference speed and GPU memory usage on different GPUs with FP16 precision.
We further validate the efficiency of our model by measuring its inference speed and memory usage on a variety of \textbf{\textit{consumer-grade} GPUs} with different memory sizes, such as 8GB (3060Ti), 11GB (2080Ti), and 24GB (4090), because the GPU environment varies for individual practitioners.
% For fair comparison, we test all models on the same GPU environment and libaries, such as CUDA, pytorch, diffusers, and so on. 
% To validate the efficiency of our model, we test our KOALA models on various \textbf{\textit{consumer-grade} GPUs}, such as 8GB (3060Ti), 11GB (2080Ti), and 24GB (4090). This ensures our model runs efficiently on lower-end GPUs and allows us to compare its performance systematically with other models across different GPU environments. 
% \cref{fig:latency_gpu} illustrates the inference speed and GPU memory usage on different GPUs with FP16 precision.
% In the 8GB GPU, all models cannot operate with FP32, and only SDM-v2.0 and KOALA-700M can run with FP32 precision. 
On the GPU-8GB, all SDXL models can't fit, while only KOALA models and SSD-Vega~\cite{gupta2024ssd} can run.
KOALA-Lightning-700M consumes comparable GPU memory but shows $2\times$ faster than SSD-Vega.
On the GPU-11GB, SDXL models can run, but KOALA-Lightning-700M still runs at approximately $5\times$ faster speed than SDXL-Base~\cite{podell2023sdxl}.
It is noted that Pixart-$\alpha$~\cite{chen2023pixartalpha} \& $-\Sigma$~\cite{chen2024pixartsigma} cannot operate on GPUs with 8GB and 11GB of memory due to their higher memory usage, but they can run on a GPU-24GB, albeit at the slowest speed.
It is worth noting that our KOALA models can operate \textit{efficiently} on all types of GPUs, highlighting the versatility and accessibility of our approach.
Furthermore, our KOALA-Lighting-700M is the best alternative for high-resolution image generation that can replace SDXL models in resource-constrained GPU environments.

% \subsection{Ablation study}

% \section{Discussion}
% \vspace{-0.2cm}
\begin{wrapfigure}{r}{0.5\textwidth}
\centering
\vspace{-0.4cm}
% \hspace{-0.1cm}
\includegraphics[width=1.\linewidth]{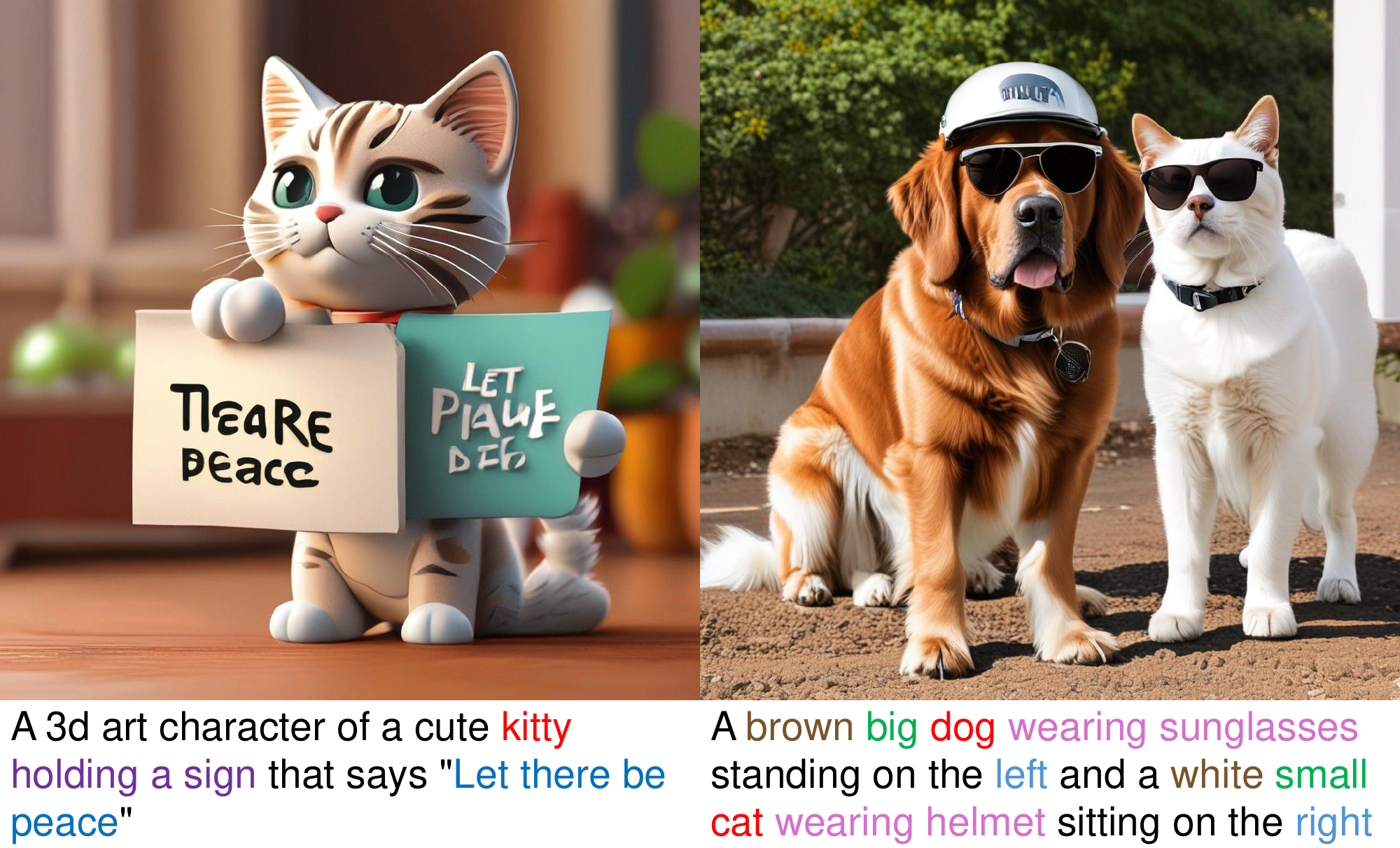}
\caption{\textbf{\footnotesize {Failure cases of KOALA-700M}}}
\vspace{-0.3cm}
\label{fig:limitation}
\end{wrapfigure}

% \begin{figure*}[h!]
% \centering
% \includegraphics[width=\textwidth]{figures/limitation.pdf}
% \caption{\textbf{Failure cases of KOALA-700M}}
% \label{fig:limitation}
% \end{figure*}

\section{Limitations}\label{sec:limitation}
While our KOALA models generate images with decent aesthetic quality, such as photo-realistic or 3D-art renderings, they still show limitations in synthesizing legible texts in the generated image as shown in~\cref{fig:limitation}~(Left).
Also, our models have difficulty in generating complex prompts with multiple attributed or object relationships, as shown in~\cref{fig:limitation}~(Right).
Additionally, since SDXL is the de facto T2I model, we have tried to compress the SDXL U-Net by addressing its bottlenecks. However, this approach is somewhat specific to the SDXL U-Net and heuristic. This limitation arises because the SDXL U-Net has a complex and heterogeneous architecture, comprising both convolutional and transformer blocks, which hinders the formulation of a more general compression principle.
More detailed investigations and examples are described in~\cref{sec:app_limitation}.

\vspace{-0.2cm}
\section{Conclusion}
% In this work, we have explored how to build memory-efficient and fast T2I models using only open-source data and models in the academic community.
In this work, we have explored how to build memory-efficient and fast T2I models, designing compact denoising U-Nets and presenting three critical lessons for boosting the performance of the efficient T2I models: 1) the importance of self-attention in knowledge distillation, 2) data characteristics, and 3) the influence of teacher models. 
Thanks to these empirical insights, our KOALA-Lightning-700M model substantially reduces the model size~(69\%$\downarrow$) and the latency~(79\%$\downarrow$) of SDXL-Base while exhibiting satisfactory generation quality.
We hope that our KOALA models can serve as cost-effective alternatives for practitioners in limited GPU environments and that our lessons benefit the open-source community in their attempts to improve the efficiency of T2I models.

% We used SDXL-Turbo~\cite{sauer2023sdxl-turbo} and SDXL-Lightning~\cite{lin2024lightning} as teacher models for feature guidance without applying their step-distillation training. For future work, since their methods and our KD approach are orthogonal, applying step-distillation to our KOALA backbone could yield synergistic effects, leading to significant speed improvements.
Additionally, since we have identified the potential of applying our self-attention-based KD to Diffusion Transformer~(DiT) models~\cite{peebles2023dit,chen2023pixartalpha,chen2024pixartsigma} in~\cref{tab:pixart} due to their architectural simplicity compared to the U-Net in SDXL, we plan to further explore more general model compression methods for DiT, as in the language model literature~\cite{gromov2024unreasonable,men2024shortgpt}, and KD techniques based on our self-attention distillation.
\section{Acknowledgments}
This work was supported by Institute of Information \& communications Technology Planning \& Evaluation (IITP) grant funded by the Korea government (MSIT) (No. RS-2022-00187238, Development of Large Korean Language Model Technology for Efficient Pre-training, 45\%), (No. 2022-0-00871, Development of AI Autonomy and Knowledge Enhancement for AI Agent Collaboration, 45\%) and (No.2019-0-00075, Artificial Intelligence Graduate School Program(KAIST), 10\%).

\bibliographystyle{template/neurips2024/icml_2024}
\bibliography{main}

\newpage
\appendix
\onecolumn
\newpage
\appendix
\onecolumn
\section*{Appendix} % 부록 시작을 나타내는 섹션
% \addcontentsline{toc}{section}{Appendix} % 부록 목차 추가
\tableofcontents % 부록의 목차
% \paragraph{Organization} The Appendix is organized as follows: First, We address KOALA's limitations in~\cref{sec:app_limitation} and describe the implementation details including training and inference in~\cref{sec:app_impl}.
% We provide 

\section{Implementation details}\label{sec:app_impl}
% \noindent
% \textbf{Training.}

\subsection{Data}\label{sec:app_data}

\noindent
\textbf{LAION-Aesthetics V2 6+}~\cite{laion-aesthetics,laion-aesthetics-6plus} includes some imperfections; thus, we conduct careful data preprocessing.
We first filtered out trivial imperfections such as blank text and corrupted images, resulting in 8,483,623 image-text pairs. 
Despite this, we observed that the text prompts in the LAION dataset are notably brief, which could hinder learning the accurate image-text correspondence. 

\begin{figure*}[ht]
\centering
\includegraphics[width=0.99\textwidth]{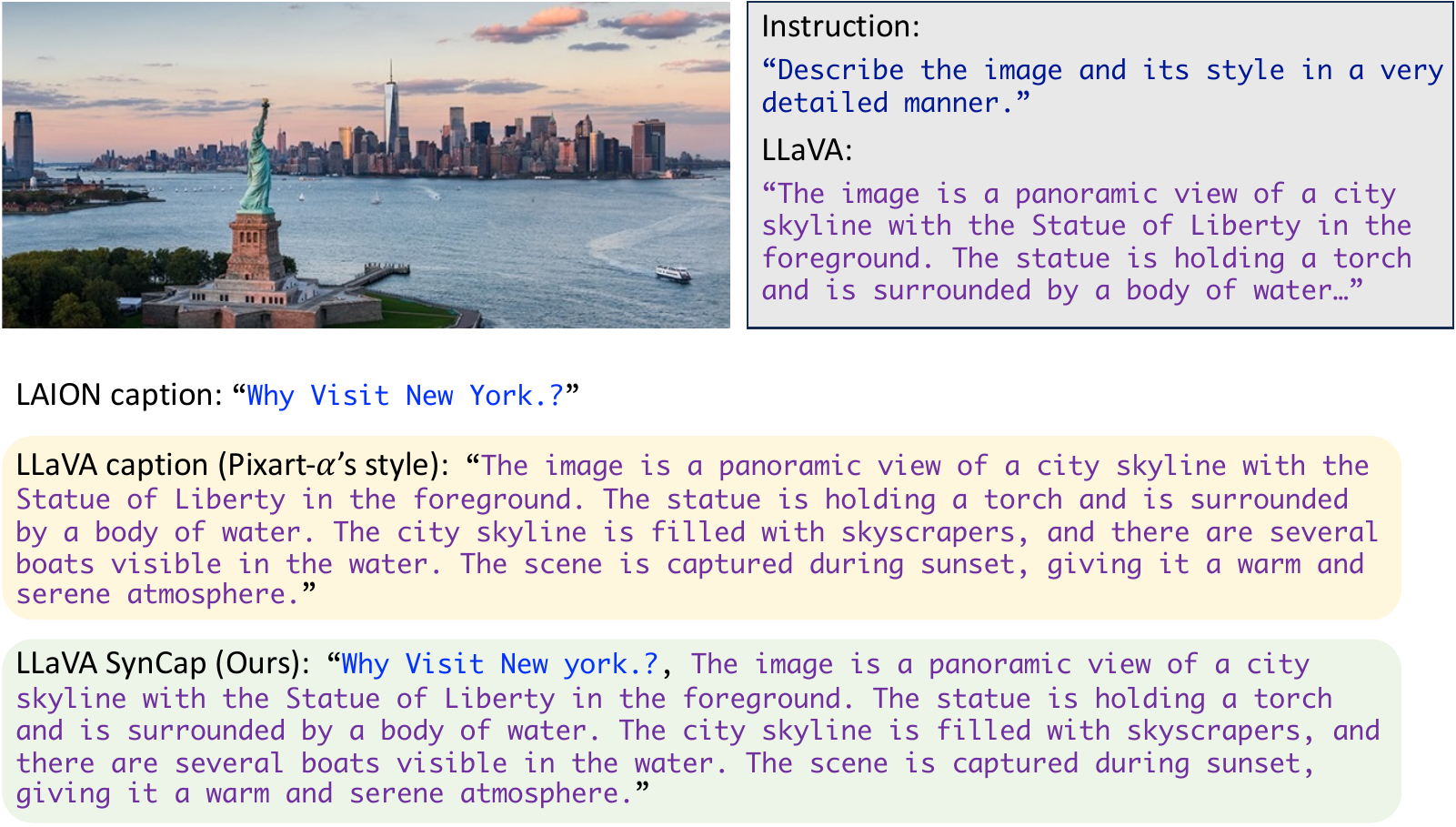}
\caption{\textbf{Sythesizing captions by LLaVA-1.5~\cite{liu2023improvedllava}.} We append the synthesized captions by LLaVA to the original ones, leveraging the existing contextual information such as proper nouns, \eg, New York
.}

\label{fig:llava_syncap}
\end{figure*}

\noindent
\textbf{Synthesized captions, \texttt{synCap}.}\label{sec:app_llava}
As some work~\cite{yu2022parti,dalle-3,chen2023pixartalpha} show that making caption data richer in information improves its generation quality, refining~(cleaning) the training caption data and synthesizing more detailed captions corresponding to each paired image by a large multimodal model, \eg, LLaVA~\cite{liu2023llava,liu2023improvedllava} would improve the image-text alignment capability.
To do this, we utilize LLaVA-1.5~\cite{liu2023improvedllava} to synthesize captions corresponding to images in LAION-Aesthetics V2 6+.
As shown in~\cref{fig:llava_syncap}, when we input the instruction to LLaVA to describe the details of the input image, we can get highly augmented captions.
It is worth noting that contrary to Pixart-$\alpha$~\cite{chen2023pixartalpha}, which replaces original captions with synthesized ones, our approach appends augmented captions to the original ones, leveraging the contextual richness of existing proper nouns~(\eg, New York).

\noindent
\textbf{LAION-POP}~\cite{LAION_POP} has a rather smaller number of images~(\eg, 491,567) but it has images with a higher resolution~($1274\times1457$) and longer average prompt length~(81) which is also generated by LMM models, CogVML~\cite{wang2023cogvlm} and LLaVA-v1.5.
We can download the dataset in the Huggingface repository\footnote{\url{https://huggingface.co/datasets/Ejafa/ye-pop}}.
We train the main models, KOALA-Turbo and KOALA-Lightning, on LAION-POP dataset in~\cref{tab:main} and~\cref{tab:main_table_full}.

\subsection{Training}\label{sec:app_training}
\noindent
\textbf{Common training protocol.}
First, we describe a common training protocol for all experiments in our work.
We base our framework on the officially released SDXL-Base-1.0\footnote{\url{https://huggingface.co/stabilityai/stable-diffusion-xl-base-1.0}} and \texttt{Diffusers} library~\cite{von-platen-etal-2022-diffusers,diffusers_sdxl}.
We mainly replace computationally burdened SDXL's U-Net with our efficient U-Net.
We keep the same two text encoders, OpenCLIP ViT-bigG~\cite{ilharco_openclip} and CLIP ViT-L~\cite{radford2021clip}, used in SDXL.
For VAE, we use \texttt{sdxl-vae-fp16-fix}~\cite{vae-fp16-fix}, which enables us to use FP16 precision for VAE computation.
We initialize the weights of our U-Net with the teacher's U-Net weights at the same block location.
We freeze the text encoders, VAE, and the teacher U-Net of SDXL and only fine-tune our U-Net.
When training, we use a discrete-time diffusion schedule~\cite{ho2020ddpm}, size- and crop-conditioning as in SDXL~\cite{podell2023sdxl}, AdamW optimizer~\cite{loshchilov2017adamw}, a batch size of 128, a constant learning rate of $10^{-5}$, and FP16 precision.

For the ablation study on the knowledge distillation strategies in~\cref{tab:distill_type,tab:distill_loc,tab:ablation}, following the common training protocol except for batch size and training iteration, for fast verification, we train our KOALA models for 30k iterations with a batch size of 32 and $1024\times1024$ resolution on LAION-Aesthetics V2 6+~\cite{laion-aesthetics-6plus} dataset using one NVIDIA A100~(80GB) GPU.

For the ablation study in Lesson 2. Data in~\cref{tab:data}, following our common training protocol, we train all cases, \eg, (a), (b), and (c) on each dataset (a) LAION-Aesthetics V2 6+ (b) LAION-Aesthetics-V2-6+ with \texttt{synCAP} and (c) LAION-POP with a batch size of 128 and $1024\times1024$ resolution for 100K iterations using 4 NVIDIA A100~(80GB) GPUs.

For the ablation study in Lesson 3. Teacher in~\cref{tab:data}, following our common training protocol, we train all cases on the LAION-POP dataset for 100K iterations using 4 NVIDIA A100~(80GB) GPUs. In particular, for using SDXL-Turbo as a Teacher model, SDXL-Turbo was originally trained with $512\times512$ resolution, so we perform KD-training using the SDXL-Turo teacher with $512\times512$ resolution.
We use the officially released checkpoint\footnote{\url{https://huggingface.co/stabilityai/sdxl-turbo}} in Hugginface.
In contrast, for using SDXL-Base and SDXL-Lightning as Teacher models, we follow their original papers with $1024\times1024$ resolution.
For the SDXL-Lightning teacher model, we use the officially released 4-step unet-checkpoint\footnote{\url{https://huggingface.co/ByteDance/SDXL-Lightning}} in Hugginface.

For the main results in~\cref{tab:main}, following our common training protocol, we finally train KOALA-Turbo and KOALA-Lightning equipped with two KOALA U-Net backbones with a batch size of 128 for 500K iterations on LAION-POP dataset using 4 NVIDIA A100~(80GB) GPUs.

For a fair comparison to our counterpart BK~\cite{kim2023bksdm} in~\cref{tab:bk}, 
we train SDM-Small proposed in BK-SDM~\cite{kim2023bksdm} with our self-attention-based KD using SDM-v1.4~\cite{sdm-v1.4} as a Teacher model, following the BK-SDM training recipe for 50K iteration with a batch size of 256 on LAION-Aesthetics V2 6.5+~\cite{laion-aesthetics-6.5plus}.
On the other hand, we train our KOALA-1B U-Net with the BK method and compare it with our KD method under the same training setup such as the same SDXL-Base-1.0 Teacher model, following the common training protocol except for 50K training iterations.

\begin{table*}[t]
\caption{
    \textbf{Quantitative comparison to state-of-the-art models} with HPSv2~\cite{wu2023hpsv2}~(Left) for \textbf{visual aesthetics} and with T2I-CompBench~\cite{huang2023compbench}~(Right) for \textbf{Image-text alignment}.
    }
\label{tab:main_table_full}
\begin{adjustbox}{width=1.01\textwidth, center}
% \begin{tabular}{@{}llccccc|ccccccc@{}}
\begin{tabular}{lrccccc|ccccccc}
\multirow{2}{*}{Model} & \multicolumn{1}{c}{\#Param.}      & \multicolumn{5}{c|}{HPSv2}                                                                                                                             & \multicolumn{3}{c}{Attribute} & \multicolumn{2}{c}{Object Relationship} & \multirow{2}{*}{Complex} & \multirow{2}{*}{Average} \\ \cmidrule(lr){3-12}
                       & \multicolumn{1}{c}{U-Net} & \multicolumn{1}{c}{Anime} & \multicolumn{1}{c}{Paintings} & \multicolumn{1}{c}{Photo} & \multicolumn{1}{c}{Concept-art} & \multicolumn{1}{c|}{Average} & Color    & Shape   & Texture  & Spatial          & Non-spatial          &                          &                          \\ \midrule
SDM-v2.0~\cite{sdm-v2.0}                   & 0.86B    & 26.34                                             & 25.41                                                 & 26.46                                             & 25.24                                                   & 25.86                                               & 0.5065                    & 0.4221                    & 0.4922                      & 0.1342                      & 0.3096                          & 0.3386                      & 0.3672                      \\
SDXL-Base-1.0~\cite{podell2023sdxl}              & 2.56B   & 32.50                                              & 30.98                                                 & 29.02                                             & 30.76                                                   & 30.82                                               & 0.6210                    & 0.5451                    & 0.5909                      & 0.1971                      & 0.3123                          & 0.4005                      & 0.4445                      \\
SDXL-Turbo~\cite{sauer2023sdxl-turbo}                 & 2.56B   & 31.48                                             & 28.17                                                 & 30.00                                                & 30.06                                                   & 29.93                                               & 0.6531                    & 0.5157                    & 0.6181                      & 0.1963                      & 0.3133                          & 0.3968                      & 0.4489                      \\
SDXL-Lightning~\cite{lin2024lightning}             & 2.56B   & 33.6                                              & 30.23                                                 & 32.42                                             & 32.48                                                   & 32.18                                               & 0.6553                    & 0.5106                    & 0.5816                      & 0.2133                      & 0.3080                          & 0.3984                      & 0.4445                      \\
Pixart-alpha~\cite{chen2023pixartalpha}               & 0.6B    & 33.45                                             & 30.80                                                  & 32.07                                             & 31.93                                                   & 32.06                                               & 0.4618                    & 0.4565                    & 0.5108                      & 0.1923                      & 0.3072                          & 0.3991                      & 0.3880                      \\
Pixart-sigma~\cite{chen2024pixartsigma}               & 0.6B    & 33.13                                             & 30.64                                                 & 31.64                                             & 31.59                                                   & 31.75                                               & 0.6107                    & 0.5463                    & 0.6172                      & 0.2538                      & 0.3091                          & 0.4302                      & 0.4612                      \\
SSD-1B~\cite{gupta2024ssd}                     & 1.3B    & 32.90                                              & 31.78                                                 & 28.87                                             & 32.18                                                   & 31.43                                               & 0.6333                    & 0.5313                    & 0.5914                      & 0.2139                      & 0.3174                          & 0.4108                      & 0.4497                      \\
SSD-Vega~\cite{gupta2024ssd}                   & 0.74B   & 33.56             & 32.53  & 29.65  & 32.95  & 32.17                                               & 0.6445                    & 0.5102                    & 0.6064                      & 0.2009                      & 0.3129                          & 0.4021                      & 0.4461                      \\ \midrule
\textbf{KOALA-Turbo-700M}           & 0.78B   & 31.03                                             & 28.57                                                 & 30.11                                             & 30.20                                                    & 29.98                                               & 0.6664                    & 0.5137                    & 0.6331                      & 0.1844                      & 0.3141                          & 0.4216                      & 0.4555                      \\
\textbf{KOALA-Turbo-1B}           & 1.16B    & 31.51                                             & 28.21                                                 & 29.80                                             & 29.85                                                   & 29.84                                               & 0.6571                    & 0.5192                    & 0.6284                      & 0.1882                      & 0.3148                          & 0.4282                      & 0.4560                      \\
\textbf{KOALA-Lightning-700M} & 0.78B   & 32.26                                             & 30.09                                                 & 31.76                                             & 31.87                                                   & 31.50                                               & 0.6605                    & 0.5179                    & 0.5953                      & 0.1969                      & 0.3102                          & 0.4223                      & 0.4505                      \\
\textbf{KOALA-Lightning-1B}   & 1.16B   & 32.52                                             & 30.54                                                 & 31.86                                             & 31.93                                                   & 31.71                                               & 0.6706                    & 0.5345                    & 0.5940                      & 0.2177                      & 0.3114                          & 0.4261                      & 0.4590                     
\end{tabular}
\end{adjustbox}
\end{table*}

\subsection{Inference}\label{sec:app_infernce}
When generating samples, we also generate images with ${1024\times1024}$ and ${512\times512}$ for KOALA-Lightning and KOALA-Turbo, FP16-precision and \texttt{sdxl-vae-fp16-fix}~\cite{vae-fp16-fix} for VAE-decoder.
Note that in the SDXL original paper~\cite{podell2023sdxl}, authors used DDIM sampler~\cite{song2020ddim} to generate samples in the figures while the diffuser's official SDXL code~\cite{sdxl_hf} used Euler discrete scheduler~\cite{karras2022elucidating} as the default scheduler. Therefore, we also use the Euler discrete scheduler for generating samples.
For KOALA-Lighting and KOALA-Turbo, we infer with 10 denoising steps.  
we set classifier-free guidance~\cite{ho2022cfg} to 3.5.
% We note that we generated samples in~\cref{fig:teaser} on NVIDIA 4090 GPUs.
For measuring latency and memory usage in fair conditions, we construct the same software environments across machines with different GPUs.
Specifically, we use \texttt{Pytorch==v2.1.2} and for a fair comparison, we don't use any speed-up tricks such as \texttt{torch.compile} and quantization.

\begin{table*}[t]
    \caption{
    \textbf{Analysis of feature level knowledge distillation of U-Net in SDXL~\cite{podell2023sdxl}.} SA, CA, and FFN denote self-attention, cross-attention, and feed-forward net in the transformer block. Res is a convolutional residual block and LF denotes the last feature~(same in BK~\cite{kim2023bksdm}). For the ablation study, we train our KOALA-1B as student U-Net for 30K iterations with a batch size of 32. 
    % We use HPSv2~\cite{wu2023hpsv2} as a visual aesthetics metric, which is more correlated with human preference than FID~\cite{heusel2017fid}.
    }
    \label{tab:ablation}
    \centering
    \begin{tabular}{c@{\hskip 0.01\textwidth}c@{\hskip 0.01\textwidth}c@{\hskip 0.01\textwidth}c}
        \hspace{-2.5cm}
        \begin{minipage}{0.18\textwidth}
            \centering
            % \scriptsize
            \fontsize{8}{10}\selectfont
            \begin{tabular}{l c}
            Distill type & HPSv2 \\ \midrule
            SD-loss & 25.53 \\ \hline
            SA	& \textbf{26.74} \\
            CA	& 26.11 \\
            Res	& 26.27 \\
            FFN	& 26.48 \\
            LF	& 26.63 \\
            \end{tabular}
            \subcaption{\textbf{Distillation type}}\label{tab:app_distill_type}
        \end{minipage} &
        \hspace{+0.25cm}
        \begin{minipage}{0.18\textwidth}
            \centering
            \fontsize{8}{10}\selectfont
            \begin{tabular}{l c}
            Distill loc. & HPSv2 \\ \midrule
            SD-loss & 25.53 \\ \hline
            DW-2	& 25.32 \\
            DW-3	& 25.57 \\
            Mid	& 25.66 \\
            UP-1	& \textbf{26.52} \\
            UP-2	& 26.05 \\
            \end{tabular}
            \subcaption{\textbf{Distill stage} }
            \label{tab:app_distill_loc}
        \end{minipage} &
        \hspace{+0.25cm}
        \begin{minipage}{0.18\textwidth}
            \centering
            \fontsize{8}{10}\selectfont
            \begin{tabular}{l c}
            SA loc.      & HPSv2 \\ \midrule
            SA-bottom	      & \textbf{26.74} \\
            SA-inter 	  & 26.58 \\
            SA-up             & 26.48 \\
            \\
            \\
            \\
            \end{tabular}
            \subcaption{\textbf{SA location.}} 
            \label{tab:app_sa_loc}
        \end{minipage} &
        \hspace{+0.25cm}
        \begin{minipage}{0.22\textwidth}
            \centering
            \fontsize{8}{10}\selectfont
            \begin{tabular}{l c}
            Combination & HPSv2 \\ \midrule
            Baseline (SA only)	& 26.74 \\ \hline
            SA + LF at DW-1 \& UP-3 	& \textbf{26.98} \\
            SA + Res at DW-1 \& UP-3 	& 26.94 \\
            SA + LF all  & 26.83 \\
            SA + Res all &	26.80 \\
            SA+CA+Res+FFN+LF all & 26.39 \\
            \end{tabular}
            % \caption{\textbf{Combination.} DW-1 \& UP-3 are the highest feature resolution in U-Net. }
            \subcaption{\textbf{Combination.}}
            \label{tab:app_combination}
        \end{minipage}
    \end{tabular}
\end{table*}

\subsection{Knowledge Distillation for Diffusion Transformer}\label{sec:app_pixart}
Following our KD strategies, we first compress Diffusion Transformer~(DiT~\cite{peebles2023dit}) backbone, DiT-XL, in Pixart-$\Sigma$~\cite{chen2024pixartsigma} by reducing the number of 28 transformers layers to 14 based on our finding in~\cref{tab:app_sa_loc}, building DiT-M.
Specifically, we select the bottom layers, \eg, from 0 to 14-th layers and remove the upper layers, \eg, from 14-th to 27-th layers.
We maintain the same embedding dimension size of 1152 for DiT-M as in DiT-XL, resulting in a model size of approximately 313M for DiT-M (compared to 611M for DiT-XL).
Then, we initialize the weights of DiT-M from the corresponding layers in DiT-XL.
For training, we optimize the same objective as ours: $\loss{task} + \loss{outKD} + \loss{featKD}$.
We conduct the ablation study in~\cref{tab:pixart} by changing the feature location from the teacher model.
Using the training recipe, codebase and \texttt{PixArt-Sigma-XL-2-512-MS} checkpoint\footnote{\url{https://huggingface.co/PixArt-alpha/PixArt-Sigma-XL-2-512-MS}} in Pixart-$\Sigma$, we train DiT-M with $512\times512$ resolution, a batch size of 192, multi-scale augmentation, CAME optimizer~\cite{luo2023came}, a constant learning rate of $2e^{-5}$, and FP16 precision for 50 epochs on LAION-POP~\cite{LAION_POP} dataset.

\subsection{Detailed formulation of training objectives}
We detail the two objectives, the $\loss{task}$ and $\loss{out}$, which are omitted in the main paper.
First, the target task loss $\loss{task}$ to learn reverse denoising process~\cite{ho2020ddpm} is summarized as:
\begin{equation}
    \loss{task} = \min_{S_\theta} \expect_{z_t, \epsilon, t, c}||\epsilon_t - \epsilon_{S_\theta}(z_t, t, c)||^2_2,
\end{equation}\label{eq:task_loss}
where $\epsilon_t$ is the ground-truth sampled Gaussian noise at timestep $t$, $c$ is text embedding as a condition, and $\epsilon_{S_\theta}(\cdot)$ denotes the predicted noise from student U-Net model, respectively.
Second, the output-level knowledge distillation (KD) loss is formulated as:
\begin{equation}
    \loss{outKD} = \min_{S_\theta} \expect_{z, \epsilon, t, c}||\epsilon_{T_\theta}(z, t, c) - \epsilon_{S_\theta}(z, t, c)||^2_2,
\end{equation}\label{eq:out_loss}
where $\epsilon_{T_\theta}(\cdot)$ denotes the predicted noise from each U-Net in the teacher model.

% \section{Full main result}\label{sec:app_main_result_full}

% \input{figures/limitation}

\begin{figure*}
\centering
\includegraphics[width=\textwidth]{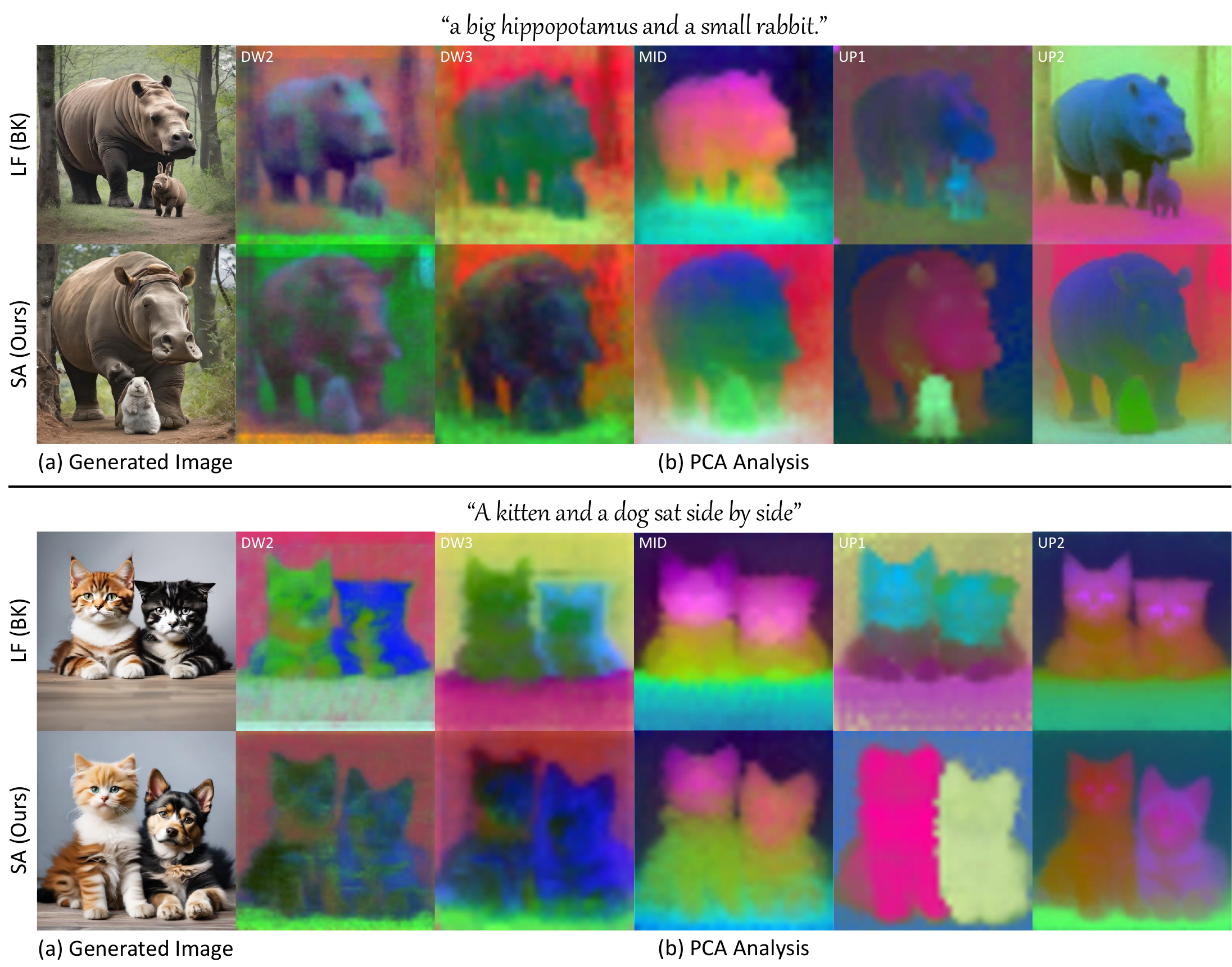}
\caption{\textbf{Extended analysis on self-attention maps of distilled student U-Nets.} (a) Generated images of LF- and SA-based distilled models, which are BK-SDM~\cite{kim2023bksdm} and our proposal, respectively. In BK-SDM's result, a rabbit or dog is depicted like a hippopotamus or cat, repectively (\ie, appearance leakage). (b) Visualization of PCA analysis results. 
% (c) Representative visualization of self-attention map from different U-Net stages. Red boxes denote the query patches.
Note that from the \texttt{UP-1} stage, the \texttt{SA}-based model \textbf{\textit{attends}} to the corresponding object (\ie, rabbit or dog) more \textit{\textbf{discriminatively}} than the \texttt{LF} model, demonstrating that self-attention-based KD allows to generate objects more distinctly.}
\label{fig:supple_attn_viz}
\end{figure*}

\section{Additional Analysis}

\subsection{Self-attention based Feature-level Knowledge distillation}\label{sec:app_kd}
In this section, we further perform analyses for how to effectively distill feature information from the Teacher model.

\noindent
\textbf{Which SA's location is effective in the transformer blocks?}
At the lowest feature level, the depth of the transformer blocks is 6 for KOALA-1B, so we need to decide which locations to distill from the 10 transformer blocks of teacher U-Net.
% in SDXL with 10 transformer blocks.
% We assume three cases; case.1 \texttt{SA-bottom}: $\sum_{i=1}^Nf^i(\cdot)$, case.2 \texttt{SA-interleave}: $\sum_{i=2n+1}^{2N}f^i(\cdot)$, and case.3 \texttt{SA-up}: $\sum_{i=10-N}^{10}f^i(\cdot)$
We assume three cases for each series of transformer blocks; (1) \texttt{SA-bottom}: $ \{ f_T^l \mid l\in\{1,2,3,4,5\}\}$, (2) \texttt{SA-interleave}: $ \{ f_T^{l} \mid l \in \{1,3,5,7,9,10\} \}$, and (3) \texttt{SA-up}: $ \{ f_T^l \mid l \in \{6,7,8,9,10\} \}$ where $l$ is the number of block.
% Please refer to the illustration in the Appendix for more details.
\cref{tab:app_sa_loc} shows that \texttt{SA-bottom} performs the best while \texttt{SA-up} performs the worst.
This result suggests that the features of the early blocks are more significant for distillation.
A more empirical analysis is described in~\cref{sec:app_cos}. 
% This result suggests that when the representation changes between blocks, the early transformers are less disparate.
Therefore, we adopt the \texttt{SA-bottom} strategy in all experiments.

\noindent
\textbf{Which combination is the best?}
In SDXL's U-Net, as shown in~\cref{fig:arch}, there are no transformer blocks at the highest feature levels (\eg, \texttt{DW-1\&UP-3}); consequently, self-attention features cannot be distilled at this stage.
Thus, we try two options: the residual block~(\texttt{Res at DW-1\&UP-3}) and the last feature~(\texttt{LF at DW-1\&UP-3}) as BK-SDM~\cite{kim2023bksdm}.
To this end, we perform SA-based feature distillation at every stage except for \texttt{DW-1} and \texttt{UP-3}, where we use the above two options, respectively.
In addition, we try additional combinations: \texttt{SA+LF all}, \texttt{SA+Res all}, and \texttt{SA+CA+Res+FFN+LF all} where \texttt{all} means all stages.
% \cref{tab:combination} demonstrates that adding more feature distillations to SA-baed distillation consistently boots performance, and especially \texttt{LF at DW1\&UP3} shows the best.
\cref{tab:app_combination} demonstrates that adding more feature distillations to the SA-absent stage~(\eg, \texttt{DW-1\&UP-3}) consistently boots performance, and especially \texttt{LF at DW1\&UP3} shows the best.
Interestingly, both \texttt{+LF all} and \texttt{+Res all} are worse than the ones at only \texttt{DW-1}\&\texttt{UP-3} and \texttt{SA+CA+Res+FFN+LF all} is also not better, demonstrating that the SA features are not complementary to the other features.

\subsection{Attention visualization for~\cref{tab:app_distill_type} and~\cref{tab:app_distill_loc}}
In Section 4.3 of the main paper, we provide empirical evidence demonstrating the paramount importance of self-attention features in the distillation process. 
% (see \cref{tab:distill_type} in the main paper)
Our findings particularly highlight the significant impact of specific self-attention (\texttt{SA}) stages (\eg, \texttt{UP-1\&UP-2}) on enhancing performance. 
To support these results, we extensively analyze self-attention maps in the main paper. 
To complete the analysis, we expand our Principal Component Analysis~\cite{jolliffe2016principal} (PCA) on self-attention maps to encompass all layers in \cref{fig:supple_attn_viz}.

As elaborated in the main paper, self-attention begins by capturing broad contextual information (\eg, \texttt{DW-2\&DW-3}) and then progressively attends to localized semantic details (\eg, \texttt{MID}). 
Within the decoder, self-attentions are increasingly aligned with higher-level semantic elements \texttt{UP-1\&UP-2}), such as objects, for facilitating a more accurate representation of appearances and structures. 
Notably, at this stage, the \texttt{SA}-based model focuses more on specific object regions than the \texttt{LF}-based model. 
This leads to a marked improvement in compositional image generation performance.

\begin{figure}[t]
\centering
\includegraphics[width=0.5\linewidth]{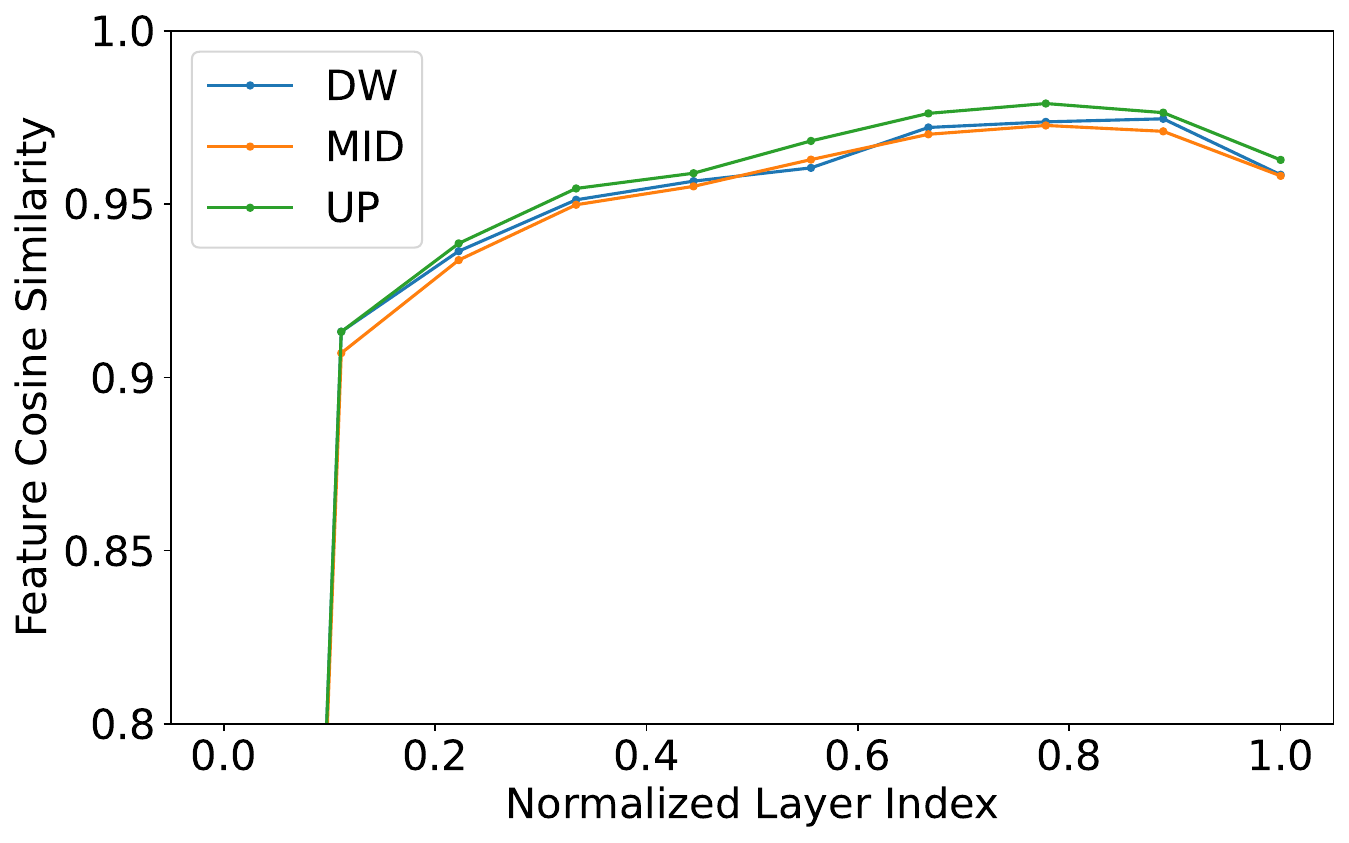}
% \vspace{-0.7cm}
\caption{\textbf{Feature cosine similarity analysis.} We plot the cross-layer cosine similarity against the normalized layer indexes of transformer block.}
\vspace{-0.5cm}
\label{fig:feat_cos}
\end{figure}

\subsection{Feature cosine similarity analysis for~\cref{tab:app_sa_loc}}\label{sec:app_cos}
KOALA models compress the computationally intensive transformer blocks in the lowest feature levels (\ie, \texttt{DW-3\&Mid\&UP-1} stages). 
Specifically, we reduce the depth of these transformer blocks from 10 to 5 for KOALA-700M and to 6 for KOALA-1B. 
For this purpose, we demonstrate that distilling knowledge from the consecutive bottom layers of transformer blocks is a simple yet effective strategy (see third finding (F3) in the main paper).

To delve deeper into the rationale behind this strategy, we conducted a thorough feature analysis of the original SDXL model~\cite{podell2023sdxl}. 
In particular, we investigate the evolution of the features within the transformer blocks.  
We compute the cross-layer cosine similarity between the output features of each block and those of its predecessors. 
A lower similarity score indicates a significant contribution of the current block, whereas a higher score implies a marginal contribution.

For this analysis, we leverage the diverse domain of prompts in the HPSv2 dataset~\cite{wu2023hpsv2}. 
We compute the cross-layer cosine similarity for each stage (\texttt{DW\&Mid\&UP}) and average these values across all prompts. 
The results are illustrated in~\cref{fig:feat_cos}. 
For all stages, transformer blocks exhibit a tendency of feature saturations: While early transformer blocks generally show a significant contribution, later blocks have less impact. 
This motivates us to distill the learned knowledge of consecutive bottom layers of transformer blocks for minimal performance degradation.

\subsection{Implementation details of \texttt{SA-bottom}}
\begin{figure}[t]
\centering
\includegraphics[width=0.5\linewidth]{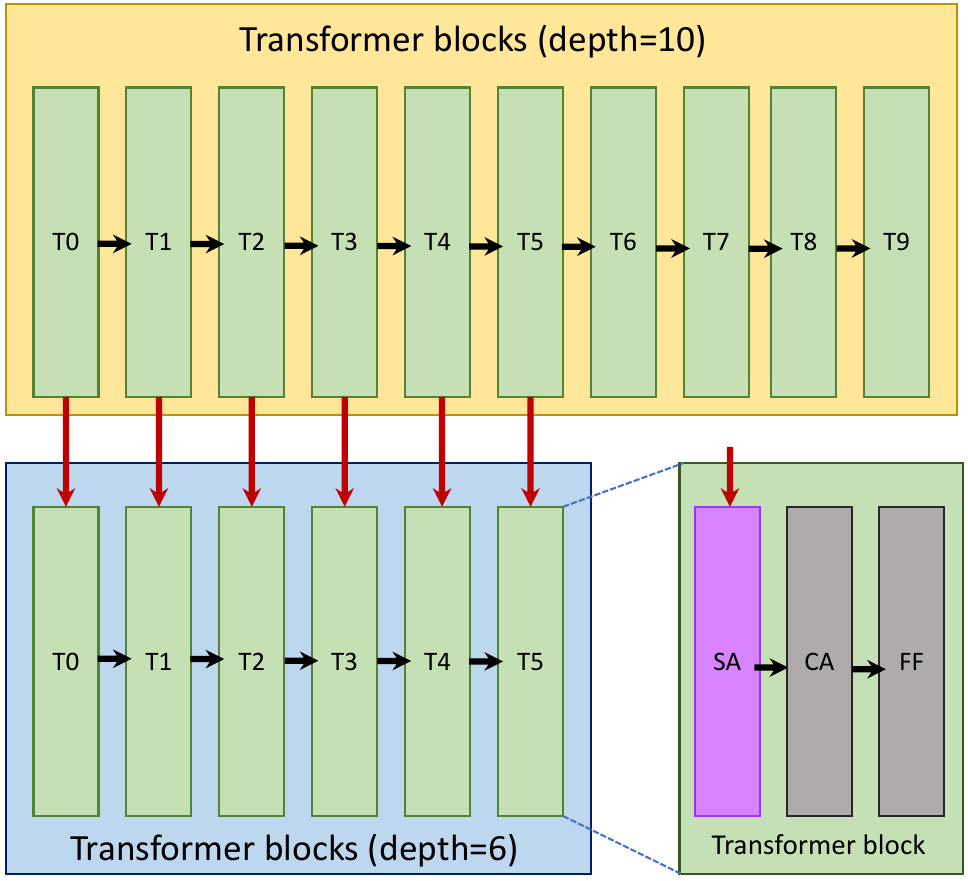}
\caption{\textbf{\texttt{SA-bottom} illustration} in~\cref{tab:app_sa_loc}.}
\label{fig:sa_bottom}
\end{figure}

\cref{fig:sa_bottom} illustrates how to choose transformer blocks when distilling self-attention~(SA) features at \texttt{DW3 \& MID \& UP1} as described in~\cref{sec:app_kd} and~\cref{tab:app_sa_loc}.
In~\cref{fig:sa_bottom}, the Transformer blocks~(yellow) with a depth of 10 is from the original SDXL's U-Net teacher model, and the Transformer blocks~(blue) with a depth of 6 is from our KOALA-1B's U-Net student model.
For \texttt{SA-bottom} in~\cref{tab:app_sa_loc}, we perform feature distillation by selecting consecutive blocks from the teacher model's transformer blocks, starting with the first one, and comparing to each transformer's self-attention~(SA) features from the student model's transformer blocks.

\begin{figure}[t]
\centering
\includegraphics[width=0.55\linewidth]{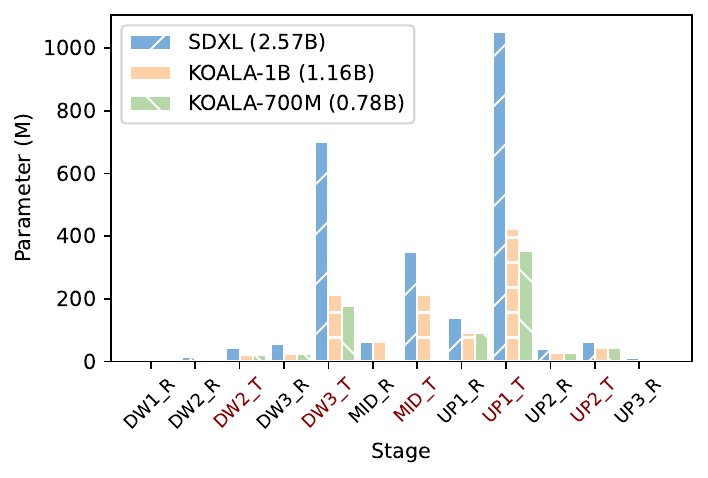}
% \vspace{-0.5cm}
\caption{\textbf{Dissection of U-Net in SDXL.} DW$i$ and UP$i$ indicate $i$-\textit{th} stage of the down and the up block, and R and T denote the Residual block and Transformer block, respectively.}
\label{fig:unet}
\end{figure}

% \subsection{Qualitative comparison for~\cref{tab:distill_type}}
\section{Qualitative results}\label{sec:app_samples}

\subsection{Represenative prompts in~\cref{fig:teaser}}\label{sec:app_prompts}
We use the following prompts for~\cref{fig:teaser}. From left-top to
right-bottom:

\begin{itemize}
    \item A 4k DSLR photo of a raccoon wearing an astronaut suit, photorealistic.
    \item A koala making latte art.
    \item A highly detailed zoomed-in digital painting of a cat dressed as a witch wearing a wizard hat in a haunted house, artstation.
    \item Cartoon of a cute hedgehog with tangled fur, standing character, looking surprised and awkward standing in a dirty puddle, dark circles under its eyes due to lack of sleep, depicted with comic exaggeration, spotlight effect highlighting its unkempt spines, use of vivid colors, high-definition digital rendering.
    \item A photorealistic render of an origami white and tan mini Bernadoodle dog standing in a surrealistic field under the moonlit setting.
    \item Peter Pan aged 60 years old, with a black background.
    \item A teddy bear wearing a sunglasses and cape is standing on the rock. DSLR photo.
    \item A photograph of a sloth wearing headphones and speaking into a high-end microphone in a recording studio.
\end{itemize}

\begin{figure*}
\centering
\includegraphics[width=1.\textwidth]{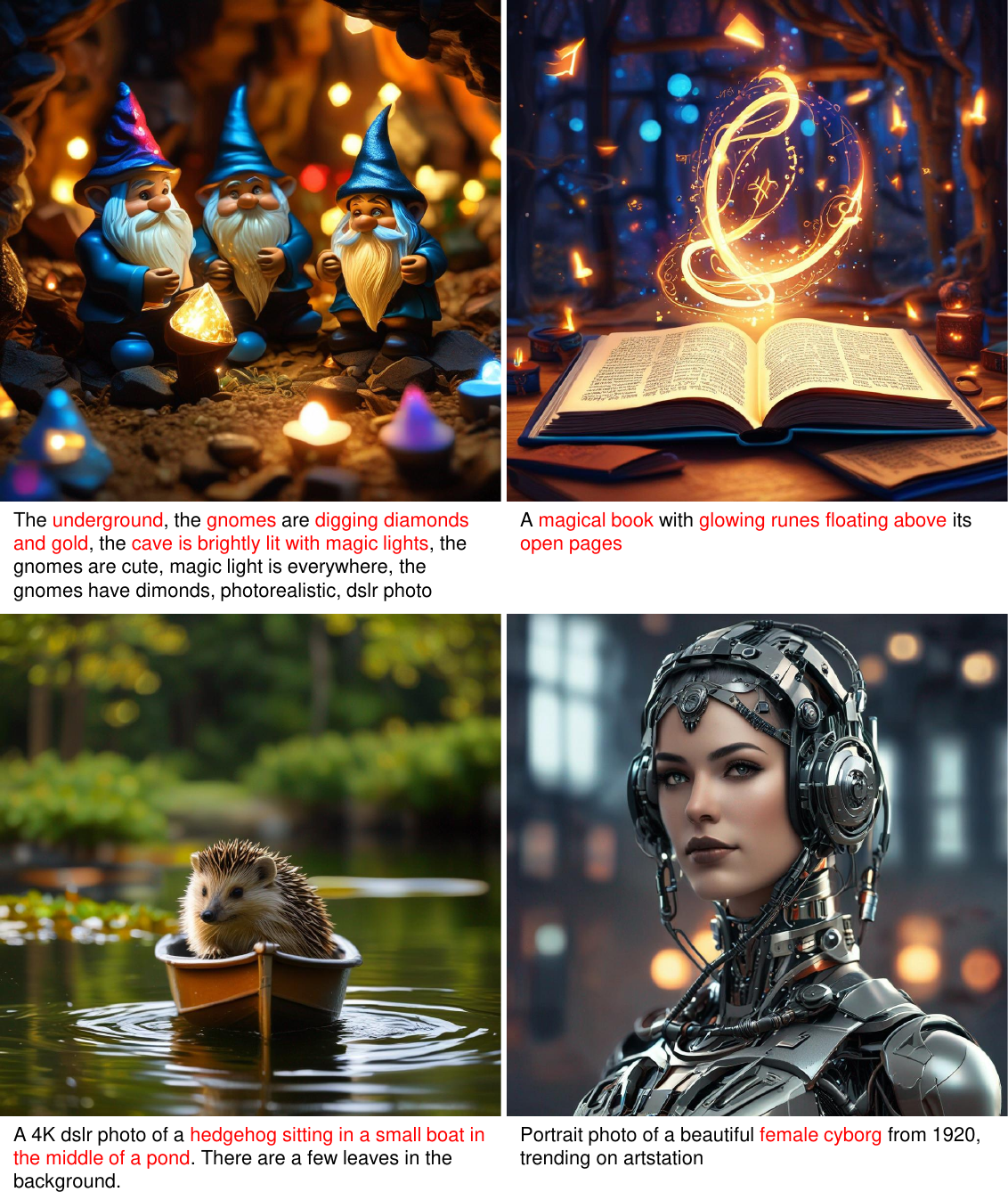}
\caption{\textbf{Qualitative results of KOALA-Lightning-1B models with $1024^2$ resolution}. 
}
\label{fig:koala_lightning_1b}
\end{figure*}

\begin{figure*}
\centering
\includegraphics[width=1.\textwidth]{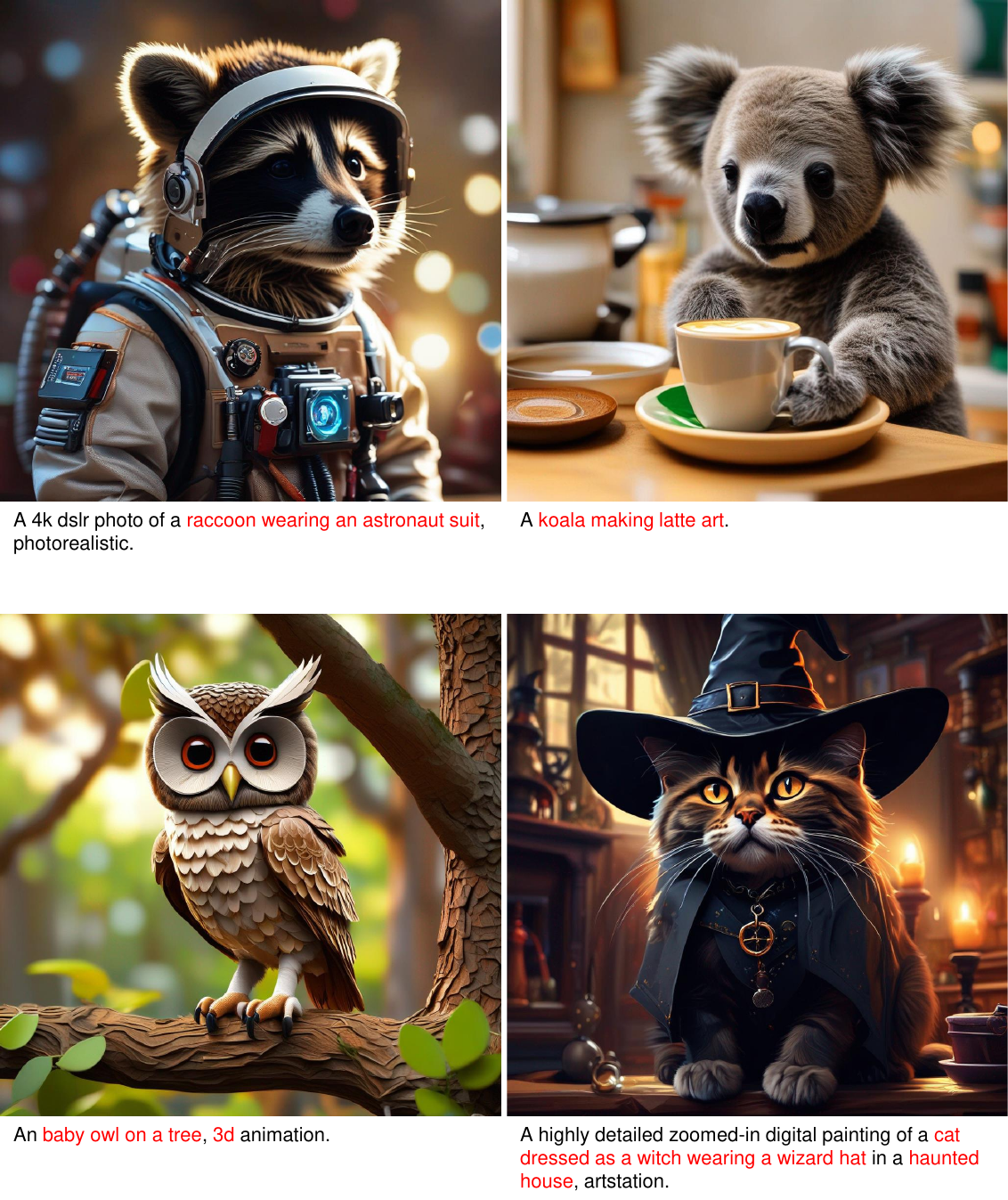}
\caption{\textbf{Qualitative results of KOALA-Lightning-1B models with $1024^2$ resolution}. 
}
\label{fig:koala_lightning_1b_2}
\end{figure*}

\begin{figure*}
\centering
\includegraphics[width=1.\textwidth]{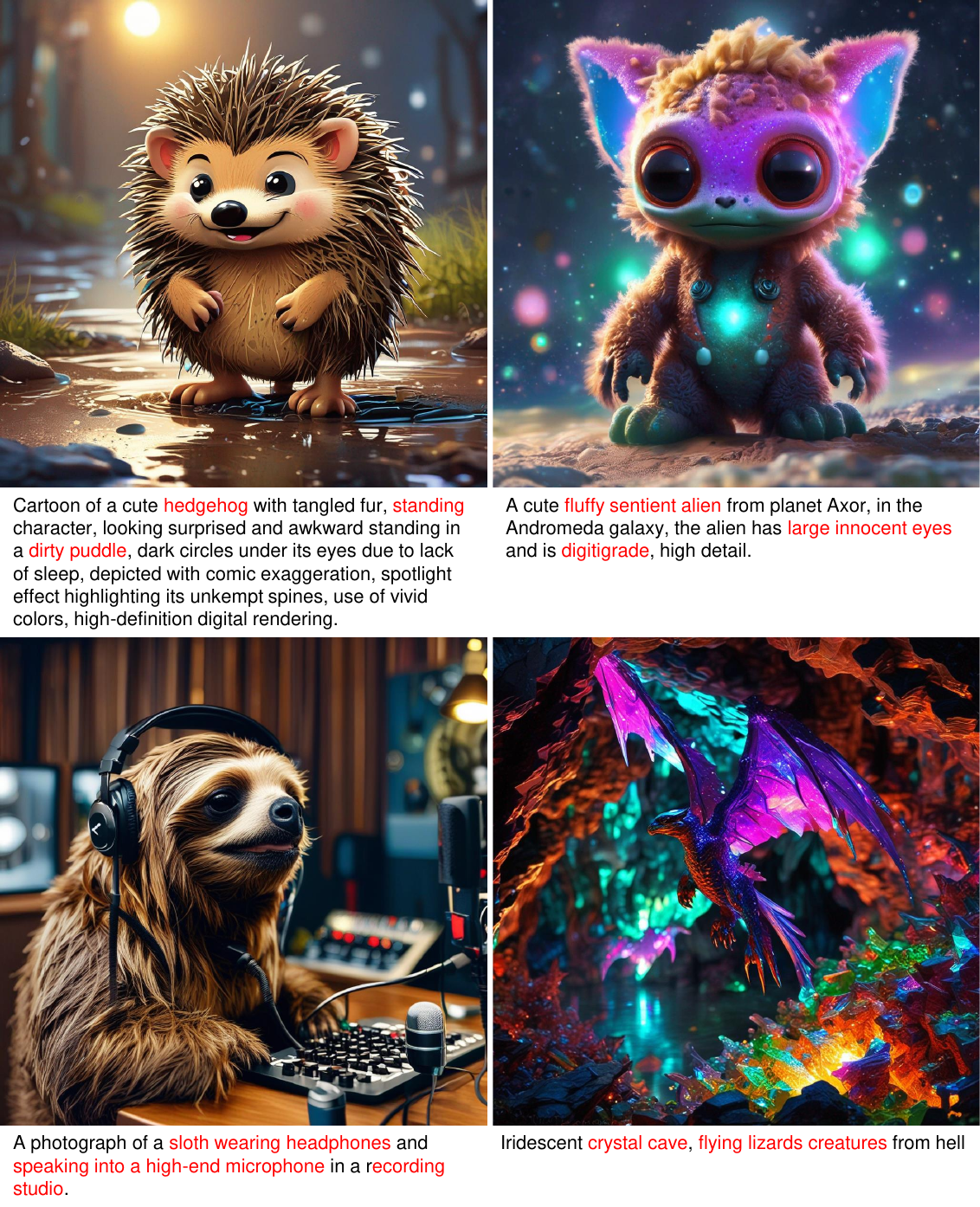}
\caption{\textbf{Qualitative results of KOALA-Lightning-1B models with $1024^2$ resolution}. 
}
\label{fig:koala_lightning_1b_3}
\end{figure*}

More qualitative results are illustrated in~\cref{fig:koala_lightning_1b,fig:koala_lightning_1b_2,fig:koala_lightning_1b_3}.

% \newpage
% \input{figures/fig_detail_light-1b}
% \input{figures/fig_detail_light-1b_2}
% \input{figures/fig_detail_light-1b_3}

% \clearpage
% \newpage
\subsection{Comparison to other methods}

\begin{figure*}
\centering
\includegraphics[width=1.\textwidth]{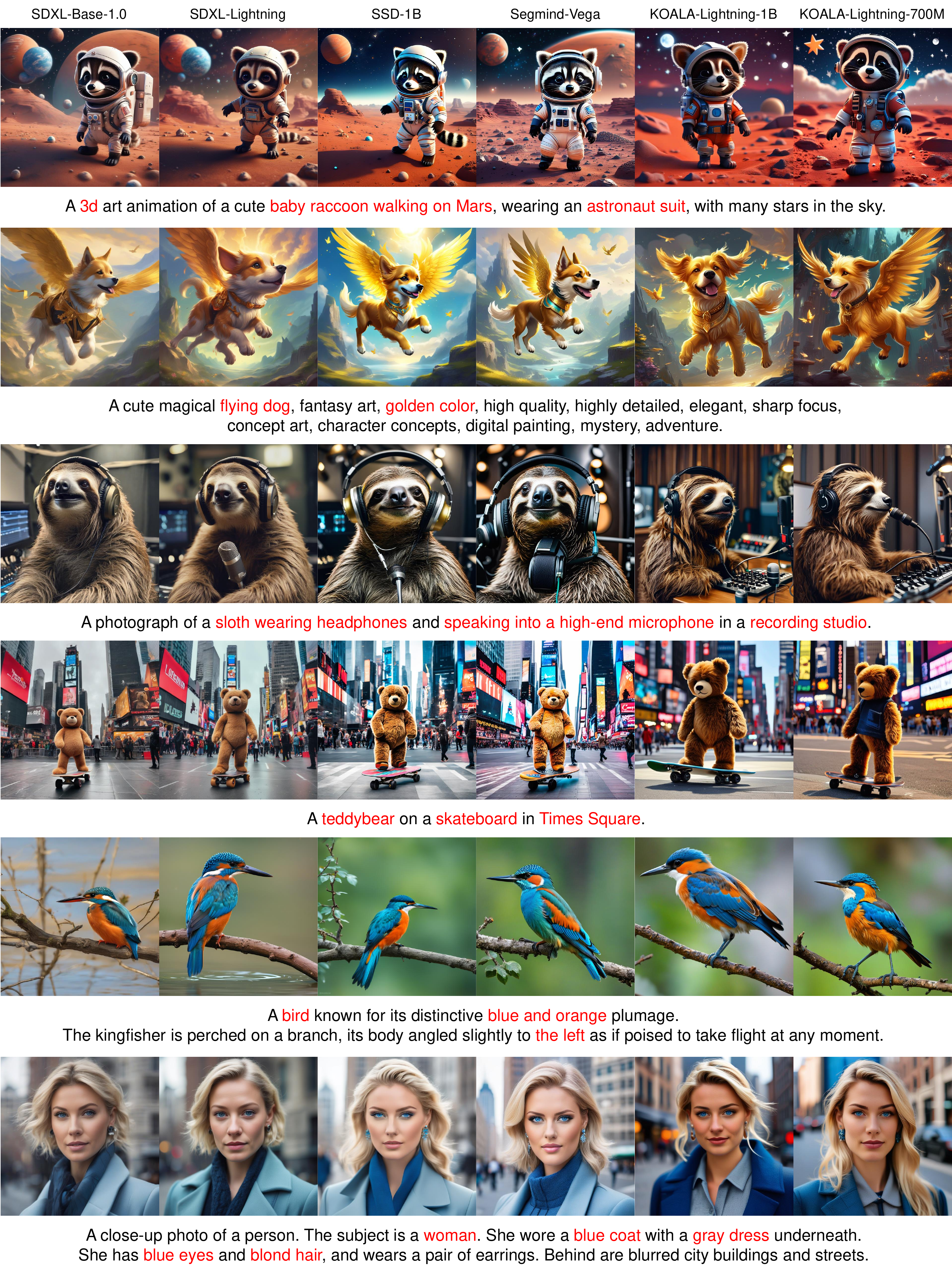}
\caption{\textbf{Qualitative comparison with state-of-the-art SDXL models: SDXL-Base-1.0~\cite{podell2023sdxl}, SDXL-Lightning~\cite{lin2024lightning}, SSD-1B~\cite{gupta2024ssd} and SSD-Vega~\cite{gupta2024ssd} with $1024^2$ resolution}. 
% 1st row: various painter styles. 
% 2nd row: various color reflections.
% 3rd row: various seasons.
% 4th row: various times.
}
\label{fig:sota_1024_comparison}
\end{figure*}

\begin{figure*}
\centering
\includegraphics[width=1.\textwidth]{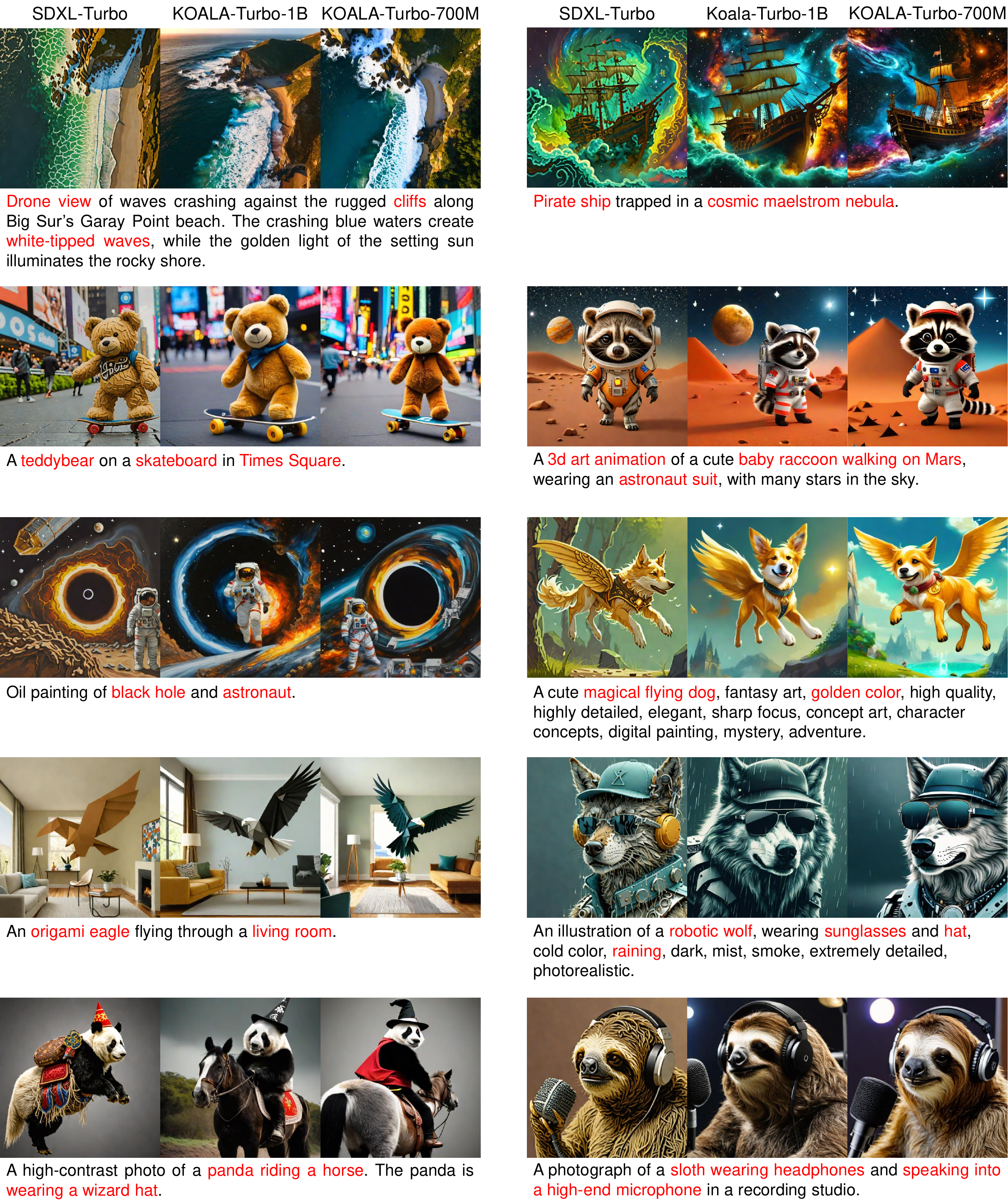}
\caption{\textbf{Qualitative comparison of SDXL-Turbo~\cite{sauer2023sdxl-turbo}} and our KOALA-Turbo models with $512^2$ resolution. }
\label{fig:sota_turbo_comparison}
\end{figure*}

We compare our KOALA-Lightning with SDXL-Base-1.0~\cite{podell2023sdxl}, SDXL-Lightning~\cite{lin2024lightning}, SSD-1B~\cite{gupta2024ssd} and SSD-Vega~\cite{gupta2024ssd} using $1024\times1024$ resolution in~\cref{fig:sota_1024_comparison}.
In addition, we compare our KOALA-Turbo with SDXL-Turo~\cite{sauer2023sdxl-turbo} using $512\times512$ resolution in~\cref{fig:sota_turbo_comparison}.

\subsection{Comparison to BK-SDM}
\begin{figure*}[h!]
\centering
\includegraphics[width=\textwidth]{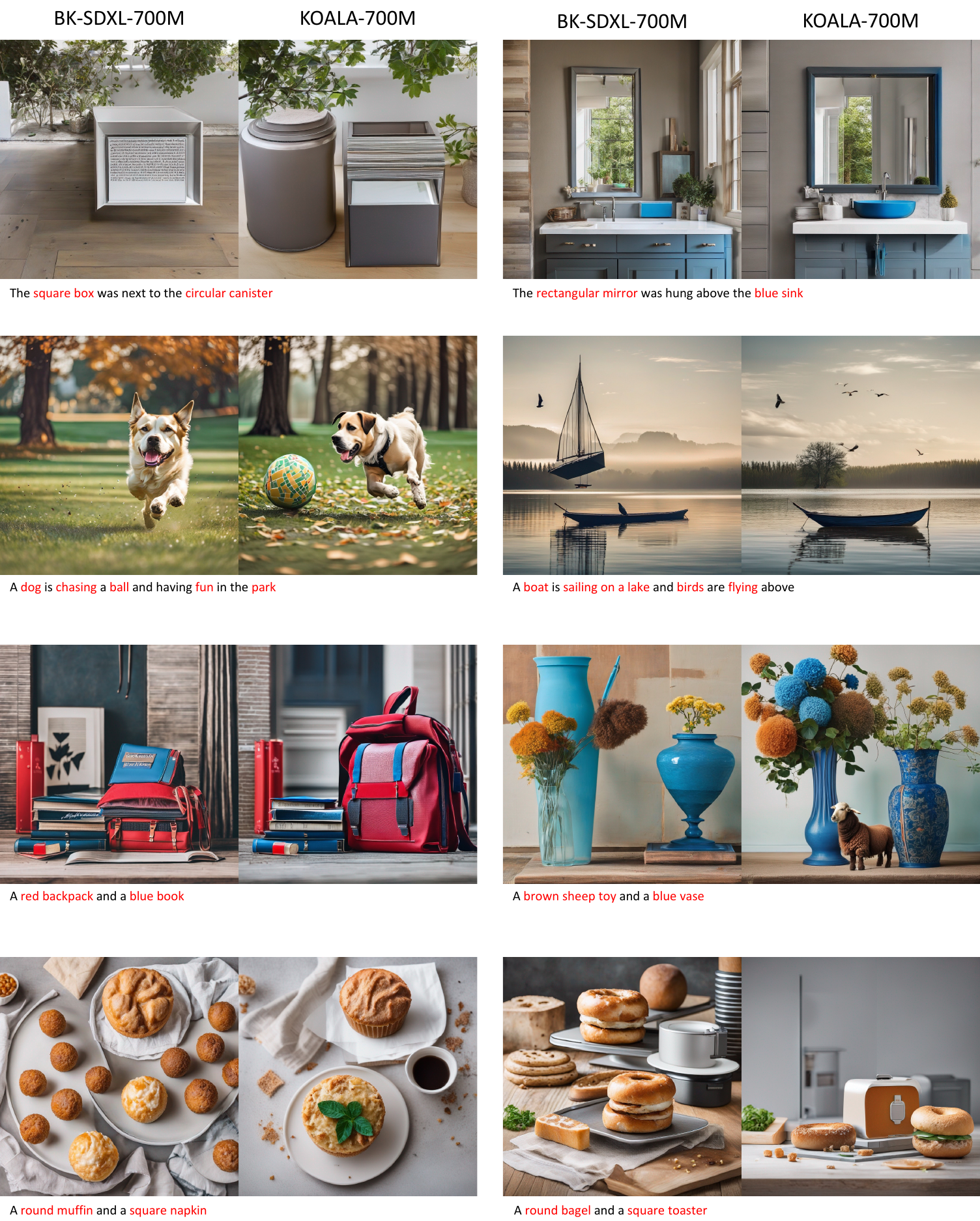}
\caption{\textbf{Qualitative comparison between BK-Base-700M vs. KOALA-700M~(ours)}. 
These models are trained with the same training recipe, such as the LAION-A+6 dataset and SDX-Base-1.0 teacher model.
% 1st row: Complex
% 2nd row: Action
% 3rd row: Color
% 4th row: Shape
}
\label{fig:bk_comparison}
\end{figure*}

In addition to the quantitative comparisons in the main paper, we also provide a qualitative comparison with BK-SDM~\cite{kim2023bksdm}. 
As illustrated in \cref{fig:bk_comparison}, BK-SDM occasionally overlooks specific attributes or objects mentioned in the text prompt and generates structurally invalid images.
On the contrary, our proposed model consistently generates images with enhanced adherence to the text, showcasing a superior ability to capture the intended details accurately.

% \section{Limitation}\label{sec:app_limitation}
\section{Failure cases}\label{sec:app_limitation}
% While our KOALA models generate images with impressive aesthetic quality~\cref{fig:teaser}, they still show limitations in several specific cases as shown in~\cref{fig:failures}:

\begin{figure*}
\centering
\includegraphics[width=1.\textwidth]{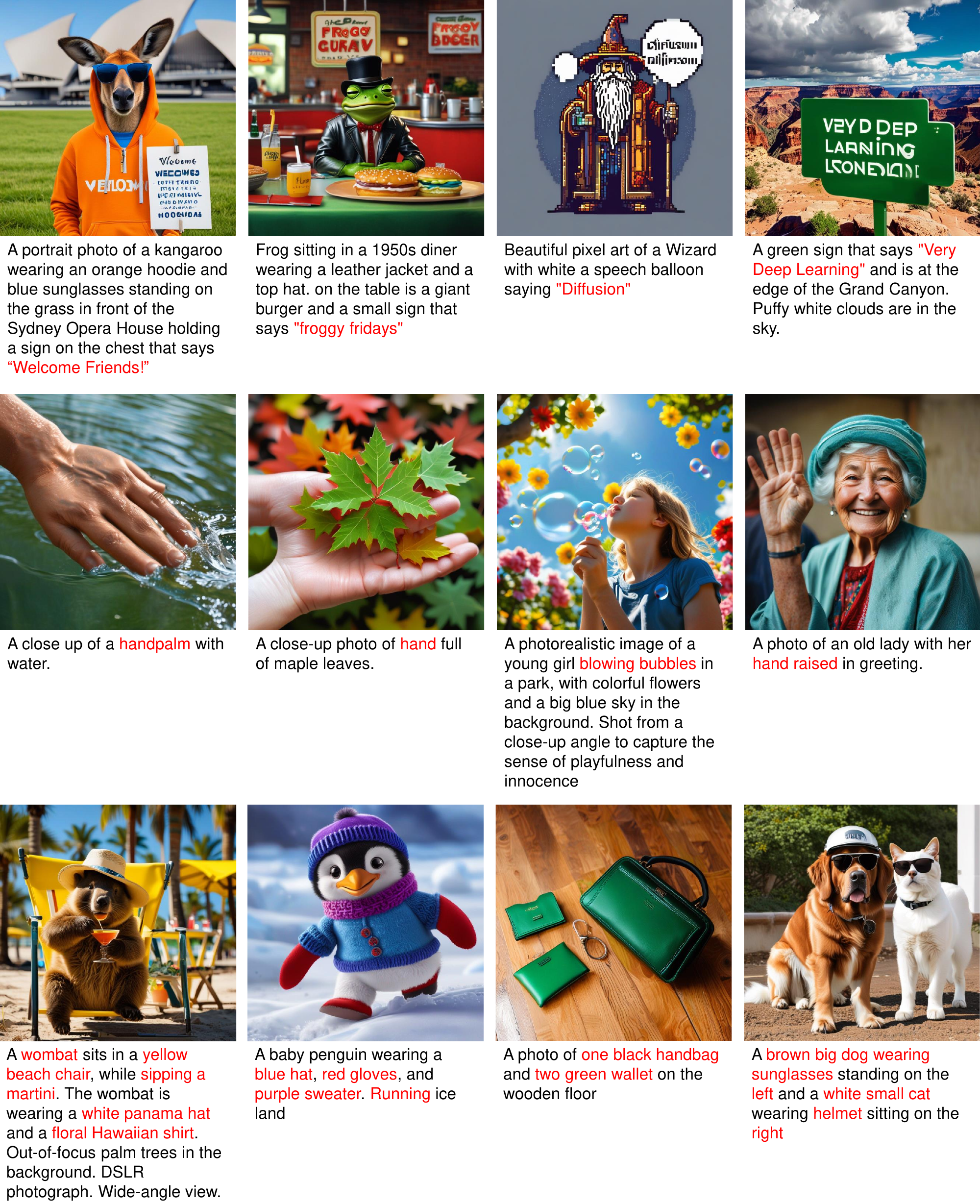}
\caption{\textbf{Failure cases of KOALA-Lightning-1B}.
% Despite its great aesthetic quality, the
KOALA-Lightning-1B model faces challenges in complex scenarios, such as rendering legible text (1st row), accurately depicting human hands (2nd row), and complex compositional prompts with multiple attributes (3rd row).
% 1st row: complex compositional prompts with multiple attributes.
% 2nd row: legible text.
% 3rd row: intricate structural details.
% 4th row: human hands.
}
\label{fig:failures}
\end{figure*}

\cref{fig:failures} illustrates that the KOALA-Lightning-1B model faces challenges in rendering legible text (the first row), accurately depicting human hands (the 2nd row), and complex compositional prompts with multiple attributes (the third row).
We conjecture that these limitations may stem from the dataset, LAION-POP dataset~\cite{LAION_POP}, we used to train, whose images don't have enough of those styles.

\noindent
\textbf{Rendering long legible text.}
We have observed that the model has difficulty in synthesizing long-legible texts in the generated image.
% Our models have difficulty in synthesizing legible texts in the generated image.
For example, as shown in ~\cref{fig:failures} (1st-row), it renders unintended letters and sometimes doesn't generate correct characters.
% For example, it renders unintended letters or generates unintelligible letters, as shown in ~\cref{fig:limitation}~(Left).
\noindent
\textbf{Complex prompt with multiple attributes.}
When attempting to compose an image using prompts that include various attributes of an object or scene, KOALA sometimes generates instances that do not perfectly follow the intended description. 
% For example, as shown in~\cref{fig:limitation}~(Right), when we configure the penguin to wear a blue hat and red gloves, only the blue hat attribute is applied, while the red gloves are not.
\textbf{human hands details.}
While we have confirmed that the model excels at representing human faces, it still struggles to render human hands. 
This may be because we haven't learned enough about the structure of the hand itself, as human hands are more often seen in conjunction with other objects or situations than in isolation.

\section{Societal Impacts}\label{sec:broad_impact}
The text-to-image generation model, our KOALA models developed in this study, has the potential to significantly advance the field of visual content creation by enabling the automated generation of diverse and creative images from textual descriptions. This innovation has numerous applications across various industries, including entertainment, education, advertising, and more. However, it is crucial to acknowledge and address the potential risks associated with the misuse of such technology, particularly concerning the generation of Not Safe For Work (NSFW) content.

To mitigate the risks associated with NSFW content, our model leverages the NSFW content detection capabilities provided by Huggingface and the transformers library. By integrating these tools, we ensure that any potentially harmful, violent, or adult content generated by KOALA is identified and filtered out before reaching the end-users. Specifically, the NSFW score is calculated for each generated image, and images with scores exceeding a predefined threshold are automatically discarded. This approach helps maintain the ethical and responsible use of our technology, promoting a safer and more positive user experience.

The adoption of such filtering mechanisms is essential to prevent the spread of inappropriate content and to adhere to ethical standards in AI development. By implementing robust NSFW detection and filtering strategies, we demonstrate our commitment to addressing broader societal concerns and promoting the responsible use of AI-generated content.

In conclusion, while the KOALA model offers significant benefits and opportunities for innovation, we recognize the importance of proactive measures to prevent its potential misuse. Our integration of NSFW content detection serves as a crucial safeguard, ensuring that our contributions to the field align with ethical guidelines and societal values.

% \section*{Broader Impact}
% Our goal of this work is to compress a large-scale text-to-image synthesis model, Stable Diffusion XL~(SDXL), by distilling the capability of SDXL into a compact model.
% Our research in the field of text-to-image generation has the potential for significant societal consequences. We acknowledge the ethical aspects and future societal implications of our work, as our model is built upon publicly available knowledge from SDXL and trained on LAION-aesthetics data. This reliance on existing data sources may introduce biases and raise concerns about the generation of toxic or harmful content. We are committed to adhering to strict ethical guidelines during data collection and processing, implementing content monitoring systems, ensuring accessibility and fairness, and actively engaging in educational and awareness efforts to minimize these potential societal impacts.

\clearpage

\begin{figure*}[h!]
\centering
\includegraphics[width=\textwidth]{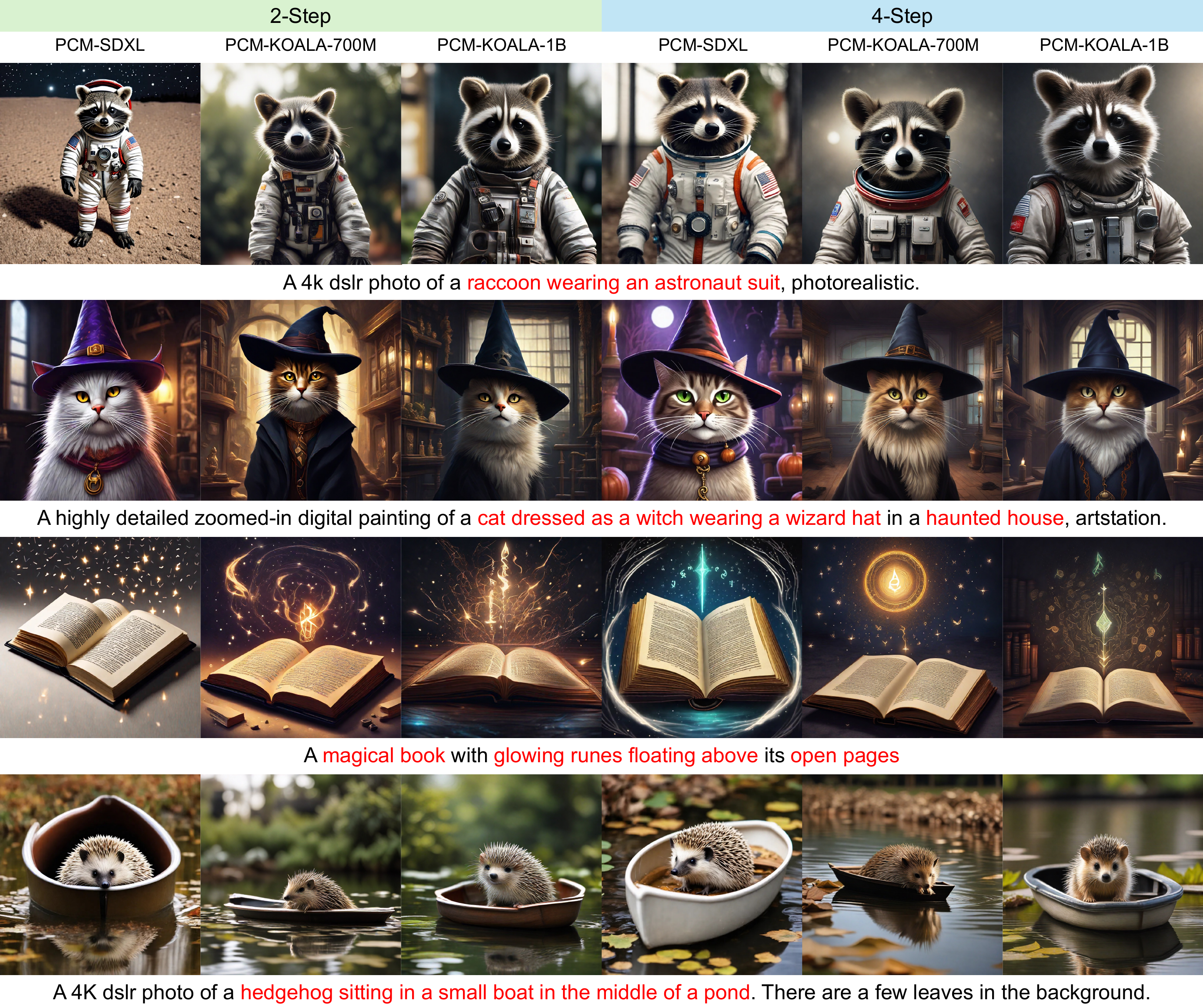}
\caption{\textbf{Qualitative comparison between PCM-SDXL and our PCM-KOALA models with $1024^2$.} 
}
\label{fig:pcm_comparison}
\end{figure*}

% % \input{figures/dreambooth}
% % \input{figures/dreambooth-1page}
% \section{Downstream task: Dreambooth}
% % The capability of personalization in text-to-image generation models is a significant topic of discussion. 
% To validate the transferability and generation capability of our KOALA model, we apply our KOALA-700M model to a custom text-to-image~(T2I) downstream task, Dreambooth~\cite{ruiz2023dreambooth}, which is a popular custom model for personalized T2I generation.
% We fine-tune our KOALA-700M model on the Dreambooth dataset using resizing 1024, the 8-bit Adam optimizer, a constant learning rate of 5e-5, and a batch size of 4 for 500 iterations without the incorporation of a class-preservation loss. The number of steps for gradient accumulation is set to 2. For generating images, we use DPM-Solver~\cite{lu2022dpm} with 25 denoising steps.
% As shown in~\cref{fig:dreambooth-1page}, with about 5-6 photographs provided, subject training is conducted alongside an identifier token, taking approximately 20 minutes per subject on an NVIDIA RTX A6000 GPU. The results demonstrate that the images are generated seamlessly, without any inconsistencies between the text and the object.2

% \newpage
% \input{sec/9_Rebuttal}

\clearpage
\newpage
\section*{NeurIPS Paper Checklist}

\begin{enumerate}

\item {\bf Claims}
    \item[] Question: Do the main claims made in the abstract and introduction accurately reflect the paper's contributions and scope?
    \item[] Answer: \answerYes{}
    \item[] Justification: We made sure that the claims made in the abstract and introduction accurately reflect our contributions and scope.

\item {\bf Limitations}
    \item[] Question: Does the paper discuss the limitations of the work performed by the authors?
    \item[] Answer: \answerYes{} % Replace by \answerYes{}, \answerNo{}, or \answerNA{}.
    \item[] Justification: We discuss the limitations of our method in \Cref{sec:limitation}, including failure cases generated by our KOALA and our simple and specific compression method.

\item {\bf Theory Assumptions and Proofs}
    \item[] Question: For each theoretical result, does the paper provide the full set of assumptions and a complete (and correct) proof?
    \item[] Answer: \answerNA{} % Replace by \answerYes{}, \answerNo{}, or \answerNA{}.
    \item[] Justification: This paper does not include theoretical results.

    \item {\bf Experimental Result Reproducibility}
    \item[] Question: Does the paper fully disclose all the information needed to reproduce the main experimental results of the paper to the extent that it affects the main claims and/or conclusions of the paper (regardless of whether the code and data are provided or not)?
    \item[] Answer: \answerYes{} % Replace by \answerYes{}, \answerNo{}, or \answerNA{}.
    \item[] Justification: We provide detailed instructions of how to reproduce all of the experiments, hyperparameters, and the models and datasets used, all of which are open-source.

\item {\bf Open access to data and code}
    \item[] Question: Does the paper provide open access to the data and code, with sufficient instructions to faithfully reproduce the main experimental results, as described in supplemental material?
    \item[] Answer: \answerNo{} % Replace by \answerYes{}, \answerNo{}, or \answerNA{}.
    \item[] Justification: We already provided where we can download the training datasets and evaluation benchmarks. However, we cannot provide the training code and checkpoint weights due to the internal policy of our company. After getting permission from our company, we will release the trained weights in Huggingface to generate images.

\item {\bf Experimental Setting/Details}
    \item[] Question: Does the paper specify all the training and test details (e.g., data splits, hyperparameters, how they were chosen, type of optimizer, etc.) necessary to understand the results?
    \item[] Answer: \answerYes{} % Replace by \answerYes{}, \answerNo{}, or \answerNA{}.
    \item[] Justification: We include details for dataset preparation and experimental settings in~\Cref{sec:app_impl}.

\item {\bf Experiment Statistical Significance}
    \item[] Question: Does the paper report error bars suitably and correctly defined or other appropriate information about the statistical significance of the experiments?
    \item[] Answer: \answerNo{} % Replace by \answerYes{}, \answerNo{}, or \answerNA{}.
    \item[] Justification:  In our study, we compared various models using quantitative evaluations, specifically focusing on HPSv2 and Compbench. The evaluation of HPSv2 required generating 2,400 images, while Compbench required 24,000 images, resulting in a significant inference cost that made repeated inferences challenging. Despite this, we believe that our performance evaluations are reliable as the performance deviations are expected to be minimal based on the sufficient number of test images used in our evaluations. Furthermore, we did not cherry-pick any images during the quantitative evaluation due to the sheer volume of images generated and evaluated. This extensive evaluation approach ensures that the reported performance metrics are representative and robust.

\item {\bf Experiments Compute Resources}
    \item[] Question: For each experiment, does the paper provide sufficient information on the computer resources (type of compute workers, memory, time of execution) needed to reproduce the experiments?
    \item[] Answer: \answerYes{} % Replace by \answerYes{}, \answerNo{}, or \answerNA{}.
    \item[] Justification: We provide the relevant details in~\cref{sec:app_impl} and~\cref{sec:gpus}.
    
\item {\bf Code Of Ethics}
    \item[] Question: Does the research conducted in the paper conform, in every respect, with the NeurIPS Code of Ethics \url{https://neurips.cc/public/EthicsGuidelines}?
    \item[] Answer: \answerYes{} % Replace by \answerYes{}, \answerNo{}, or \answerNA{}.
    \item[] Justification: We have thoroughly reviewed the ethics guidelines and we have made sure to preserve anonymity for the review process.

\item {\bf Broader Impacts}
    \item[] Question: Does the paper discuss both potential positive societal impacts and negative societal impacts of the work performed?
    \item[] Answer: \answerYes{} % Replace by \answerYes{}, \answerNo{}, or \answerNA{}.
    \item[] Justification: We already addressed the broader impacts in~\cref{sec:broad_impact}.
    
\item {\bf Safeguards}
    \item[] Question: Does the paper describe safeguards that have been put in place for responsible release of data or models that have a high risk for misuse (e.g., pretrained language models, image generators, or scraped datasets)?
    \item[] Answer: \answerYes{} % Replace by \answerYes{}, \answerNo{}, or \answerNA{}.
    \item[] Justification: As a text-to-image generation model, our KOALA, addresses the potential risks of generating Not Safe For Work (NSFW) content, which includes harmful, violent, or adult imagery. To mitigate these risks, KOALA integrates NSFW content detection capabilities provided by Huggingface and the transformers library. By calculating the NSFW score for each generated image and filtering out those that exceed a predefined threshold, our model effectively prevents the creation and dissemination of inappropriate content. This approach ensures the ethical and responsible use of AI technology, aligning our contributions with societal values and ethical guidelines.

\item {\bf Licenses for existing assets}
    \item[] Question: Are the creators or original owners of assets (e.g., code, data, models), used in the paper, properly credited and are the license and terms of use explicitly mentioned and properly respected?
    \item[] Answer: \answerYes{} % Replace by \answerYes{}, \answerNo{}, or \answerNA{}.
    \item[] Justification: we have clearly cited the teacher models used for training KOALA: SDXL-Base, SDXL-Turbo, and SDXL-Lightning in the main paper and Appendix.

\item {\bf New Assets}
    \item[] Question: Are new assets introduced in the paper well documented and is the documentation provided alongside the assets?
    \item[] Answer: \answerNA{} % Replace by \answerYes{}, \answerNo{}, or \answerNA{}.
    \item[] Justification: We didn't provide new assets except for our main paper.

\item {\bf Crowdsourcing and Research with Human Subjects}
    \item[] Question: For crowdsourcing experiments and research with human subjects, does the paper include the full text of instructions given to participants and screenshots, if applicable, as well as details about compensation (if any)? 
    \item[] Answer: \answerNA{} % Replace by \answerYes{}, \answerNo{}, or \answerNA{}.
    \item[] Justification: We have not used crowdsourcing or human subjects.

\item {\bf Institutional Review Board (IRB) Approvals or Equivalent for Research with Human Subjects}
    \item[] Question: Does the paper describe potential risks incurred by study participants, whether such risks were disclosed to the subjects, and whether Institutional Review Board (IRB) approvals (or an equivalent approval/review based on the requirements of your country or institution) were obtained?
    \item[] Answer: \answerNA{} % Replace by \answerYes{}, \answerNo{}, or \answerNA{}.
    \item[] Justification: We have not used crowdsourcing or human subjects.

\end{enumerate}

\end{document}